\documentclass[11pt]{article}

\usepackage{acl}

\usepackage{times}
\usepackage{latexsym}
\usepackage[T1]{fontenc}
\usepackage[utf8]{inputenc}
\usepackage{microtype}
\usepackage{inconsolata}
\usepackage{graphicx}
\usepackage{multirow}

\usepackage{amsmath}
\usepackage{amssymb}
\usepackage{amsthm}
\usepackage{booktabs}
\usepackage{algorithm}
\usepackage{algorithmic}

\title{Variance-Adaptive Muon: Pre-Orthogonalization Variance Modulation for Efficient Language Model Pretraining}

\author{
Jingru Li, Yibo Fan, Huan Li\thanks{Corresponding author.} \\
College of Artificial Intelligence, Nankai University, Tianjin, China \\
\texttt{\{jingru\_lee,yibofan\}@mail.nankai.edu.cn}, \texttt{lihuanss@nankai.edu.cn}
}

\begin{document}
\maketitle
\begin{abstract}

Optimizer design plays a central role in efficient language model pretraining, directly affecting optimization dynamics, convergence speed, and compute cost under fixed training budgets.
Muon has emerged as a strong optimizer by orthogonalizing momentum updates, yielding a matrix-valued analogue of sign-based normalization.
However, unlike Adam-style methods, Muon does not explicitly incorporate gradient-variance information into its updates.
Motivated by Adam's variance-adaptive interpretation, we propose Muon-NSR and Muon-VS, two variance-adaptive Muon variants for language model pretraining.
Muon-NSR applies noise-to-signal ratio (NSR) modulation before Newton--Schulz orthogonalization, whereas Muon-VS uses variance scaling (VS) without introducing any additional hyperparameters beyond those of Muon.
Both methods preserve Muon's spectral normalization structure while requiring only one additional variance buffer.
Experiments on Llama-style and GPT-2 pretraining across model scales from 125M to 1.2B parameters show that our methods improve over well-tuned Muon baselines and remain competitive with representative adaptive Muon-family baselines.
On Llama-1.2B, Muon-VS achieves a 1.33$\times$ step-to-target speedup over a well-tuned Muon baseline, with Muon's final validation loss as the target.
These results indicate that variance-adaptive modulation is a simple and effective mechanism for improving Muon-style optimizers in language model pretraining.

\end{abstract}

\begin{figure}[t]
\centering
\includegraphics[width=\columnwidth]{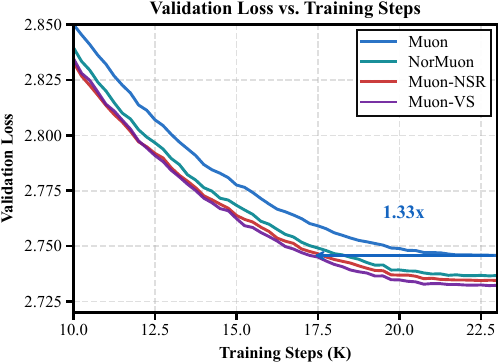}
\caption{\textbf{Suite~A Llama-1.2B pretraining.}
Validation loss on the C4-EN validation split during pretraining on DCLM with the $1\times$ Chinchilla token budget.}
\label{fig:llm_stf_1.2B_val_loss_vs_iter}
\end{figure}

\section{Introduction}

Large language models (LLMs) are commonly developed through self-supervised pretraining on large-scale text corpora, a paradigm that has become central to modern natural language processing (NLP)~\citep{raffel2020exploring,brown2020language}.
Scaling studies show that this paradigm incurs substantial computational cost that grows jointly with model size and the total number of training tokens~\citep{kaplan2020scaling,hoffmann2022empirical}.
Under a fixed architecture and data budget, optimizer design provides a direct algorithmic lever for improving pretraining efficiency by reducing the number of steps or wall-clock time required to reach a target validation loss~\citep{wen2025fantastic,semenov2025benchmarking,shah2025practical}.


Adam~\citep{kingma2014adam} and its variants, particularly AdamW~\citep{loshchilov2017decoupled}, remain standard choices for Transformer-based language model pretraining~\citep{wen2025fantastic,liang2024cautious}.
Recent analysis shows that when the first- and second-moment exponential moving averages (EMAs) use the same decay rate, Adam can be interpreted as a variance-adaptive sign update that attenuates high-variance coordinates~\citep{orvieto2026search}.
Under this interpretation, Adam's per-coordinate update decomposes into two complementary mechanisms: \emph{sign-based coordinate-wise normalization} and \emph{variance-based noise-aware damping}.

Muon~\citep{jordan2024muon} extends sign normalization to matrix-valued momentum updates by approximating the orthogonal polar factor of a momentum matrix with Newton--Schulz iterations, yielding a spectrally normalized analogue of a sign update.
This orthogonalized update has made Muon competitive in language model pretraining and optimizer benchmarks~\citep{wen2025fantastic,liu2025muon,shah2025practical}, and industry-scale reports further associate Muon with faster convergence and improved training stability~\citep{deepseekai2026deepseekv4}.
Despite these empirical strengths, standard Muon captures the spectral analogue of sign normalization but lacks online variance statistics and thus explicit variance-aware damping of high-variance or noise-dominated update directions.
This mechanistic gap therefore motivates the central question.

\emph{Can Adam-style variance-adaptive modulation be incorporated into Muon's matrix-valued sign updates for efficient language model pretraining?}

\paragraph{Contributions.}
To address this question while preserving Muon's spectral normalization, we propose \textbf{Muon-NSR} and \textbf{Muon-VS}, two variance-adaptive Muon variants for efficient language model pretraining. Our contributions are twofold:

\begin{itemize}
    \item \textbf{Variance-adaptive Muon updates.}
    Using online centered gradient-variance estimates, Muon-NSR attenuates high-uncertainty update entries via noise-to-signal ratio (NSR) modulation, while Muon-VS applies variance scaling without introducing additional optimizer-specific hyperparameters.
    Both variants modulate the update proxy before Newton--Schulz orthogonalization, thereby preserving Muon's spectral update structure with only one additional variance buffer.

    \item \textbf{Matched pretraining evaluation.}
    We evaluate Muon-NSR and Muon-VS across Llama-style and GPT-2 pretraining regimes spanning model scales from 125M to 1.2B parameters.
    Under matched training recipes and fixed pretraining budgets, both variants converge faster and achieve lower validation loss than the evaluated AdamW and Muon-family baselines.
    On Llama-1.2B under the Suite~A protocol~\citep{wen2025fantastic}, Muon-VS achieves a 1.33$\times$ step-to-target speedup over Muon, using Muon's final validation loss as the target.
\end{itemize}

\section{Related Work}

Stochastic gradient descent~\citep{robbins1951stochastic} and its momentum variants~\citep{polyak1964some} are classical first-order optimization methods.
Sign-based coordinate-wise normalization discards update magnitudes while preserving directional information.
signSGD~\citep{bernstein2018signsgd} applies this normalization to stochastic gradients, while Signum~\citep{bernstein2018signsgd} and Lion~\citep{chen2023symbolic} extend it to momentum directions.

Beyond sign-based normalization, adaptive gradient methods rescale coordinates using gradient statistics, including AdaGrad~\citep{duchi2011adaptive} and RMSProp~\citep{tieleman2012lecture}.
Adam~\citep{kingma2014adam} combines momentum with second-moment adaptivity.
Moreover, under tied first- and second-moment EMA decay rates, Adam can be interpreted as a sign update with variance-based damping~\citep{orvieto2026search}.
AdamW~\citep{loshchilov2017decoupled} further decouples weight decay from Adam's update.

Coordinate-wise adaptive methods commonly view model parameters as a vector and apply diagonal rescaling per coordinate, without explicitly exploiting layer-wise parameter-block structures such as vectors and matrices~\citep{lau2025polargrad}.
K-FAC~\citep{martens2015optimizing} approximates natural-gradient updates using Kronecker-factored curvature, whereas Shampoo~\citep{gupta2018shampoo} and SOAP~\citep{vyas2024soap} use structured matrix preconditioning.
Muon~\citep{jordan2024muon} instead approximates the orthogonal polar factor of a momentum matrix with Newton--Schulz iterations, yielding a matrix-valued analogue of sign-based coordinate-wise normalization that maps nonzero singular values approximately to one.

Muon's success has motivated extensions incorporating adaptive normalization into its spectral-update framework.
AdaMuon~\citep{si2025adamuon} combines element-wise second-moment normalization after orthogonalization with sign-stabilized orthogonal updates.
NorMuon~\citep{li2025normuon} applies row-wise normalization to orthogonalized updates using neuron-wise second moments.
Muon$^2$~\citep{liu2026muon} preconditions Muon's momentum using Adam-style uncentered second moments before Newton--Schulz orthogonalization.
By contrast, we use centered variance estimates, rather than uncentered second moments or post-orthogonalization statistics, to introduce Adam-inspired noise-aware damping before orthogonalization.

\section{Motivation from the Variance-Adaptive Perspective of Adam}
\label{sec:prelim}

\paragraph{An equivalent form of Adam with equal betas.}

Consider a scalar coordinate $x_t \in \mathbb{R}$ at iteration $t$. Let $g_t \in \mathbb{R}$ be the stochastic gradient sample, $m_t$ and $v_t$ be the EMAs of the first and second moments, respectively, $\beta_1,\beta_2 \in (0,1)$ be their decay rates, $\eta_t>0$ be the learning rate, and $d_t$ be the coordinate-wise preconditioned update direction. For matrix- or vector-valued quantities, unless stated otherwise, powers, square roots, and divisions are understood element-wise. Omitting bias correction and the stability term $\varepsilon$ for analytical tractability, the Adam update can be written as
\begin{align}
&m_t=\beta_1 m_{t-1}+(1-\beta_1) g_t,\label{adam1}\\
&v_t=\beta_2 v_{t-1}+(1-\beta_2) g_t^2,\label{adam2}\\
&d_t=\frac{m_t}{\sqrt{v_t}}=\frac{m_t}{\sqrt{m_t^2+v_t-m_t^2}},\label{adam3}\\
&x_{t+1}=x_t-\eta_t d_t.\notag
\end{align}
Recently, \citet{orvieto2026search} observed that employing a shared EMA decay rate for both first- and second-moment estimators, denoted as $\beta_1=\beta_2=\beta$, maintains competitive performance across various settings. Under this shared parameterization, \citet[Proposition~1]{orvieto2026search} proved that $v_t-m_t^2$ satisfies the recursion
\begin{equation}
\label{eq:reg-adam-ml}
\begin{aligned}
v_t - m_t^2
&= \beta\left(v_{t-1} - m_{t-1}^2\right) \\
&\quad + \beta(1-\beta)\left(m_{t-1} - g_t\right)^2 .
\end{aligned}
\end{equation}
Replacing \eqref{adam2} with \eqref{eq:reg-adam-ml} yields an equivalent form.

\paragraph{Interpretation of Adam as a variance-adaptive sign update.}
Assume that the stochastic gradient $g_t$ at iteration $t$ is generated from a Gaussian distribution with unknown and slowly varying mean and variance,
\(
g_t \sim \mathcal{N}(\mu,\sigma^2).
\)
Let $(\mu_{t-1},\sigma_{t-1}^2)$ denote the estimate of the gradient's mean and variance from the previous iteration. Upon observing $g_t$, the estimate $(\mu_t,\sigma_t^2)$ is defined as the solution of the following regularized maximum likelihood estimation problem~\citep{orvieto2026search}:
\begin{align}
(\mu_t, \sigma_t^2)
&= \operatorname*{arg\,min}_{\mu \in \mathbb{R},\, \sigma^2 > 0}
\Bigl\{ -\log p(g_t \mid \mu, \sigma^2) \notag \\
&\hspace{-1.5em}+ \frac{1}{\lambda}\,
\operatorname{KL}\!\bigl(
\mathcal{N}(\mu_{t-1}, \sigma_{t-1}^2)
\,\|\, 
\mathcal{N}(\mu, \sigma^2)
\bigr) \Bigr\},
\label{eq:reg-ml}\raisetag{1.0\baselineskip}
\end{align}
where $\lambda > 0$ controls the trade-off between fitting the current observation and penalizing deviation from the previous estimate.
The standard closed forms of the Gaussian negative log-likelihood and the Gaussian--Gaussian Kullback--Leibler divergence used here are omitted for brevity and follow those in \citet{orvieto2026search}.

With the identification $\beta = \frac{1}{1+\lambda} \in (0,1)$, \citet[Theorem 4.1]{orvieto2026search} proved that the minimizer of~\eqref{eq:reg-ml} admits a closed-form solution that is recursively defined as follows:
\begin{align}
    \mu_t &= \beta \mu_{t-1} + (1-\beta)g_t,
    \label{eq:mk-update}\\
    \sigma_t^2 &= \beta\sigma_{t-1}^2 + \beta(1-\beta)(\mu_{t-1}-g_t)^2.
    \label{eq:sig-update}
\end{align}
Comparing \eqref{eq:reg-adam-ml} with \eqref{eq:sig-update} and \eqref{adam1} with \eqref{eq:mk-update} reveals that $\sigma_t^2=v_t-m_t^2$ and $\mu_t=m_t$, implying that $m_t$ and $v_t-m_t^2$ in Adam's updates serve as online estimators of the local gradient mean and variance, respectively.
Substituting these identities into \eqref{adam3} yields the variance-adaptive update
\begin{equation}
    d_t 
    = 
    \frac{m_t}{\sqrt{m_t^2+\sigma_t^2}} 
    = 
    \frac{\operatorname{sign}(m_t)}{\sqrt{1+\sigma_t^2/m_t^2}}, 
    \quad (m_t \neq 0),
    \label{eq:variance-scaled}
\end{equation}
where $\mathrm{NSR}_t=\sigma_t/|m_t|$ denotes the estimated local noise-to-signal ratio.
Under the shared-decay interpretation, the update in \eqref{eq:variance-scaled} decomposes Adam's per-coordinate update into two coupled mechanisms.
The term $\operatorname{sign}(m_t)$ corresponds to \emph{sign-based coordinate-wise normalization}~\citep{bernstein2018signsgd}.
The factor $(1+\mathrm{NSR}_t^2)^{-1/2}$ corresponds to \emph{variance-based noise-aware damping}, suppressing step sizes in noise-dominated regimes~\citep{orvieto2026search}.

\paragraph{Muon's Spectral Sign Update.}
Muon~\citep{jordan2024muon} applies the matrix sign map, defined as the orthogonal polar factor, to a Nesterov-style momentum update proxy.
This operation generalizes sign-based normalization from scalar coordinates to matrices by replacing coordinate-wise sign normalization with spectral normalization.
For a matrix-valued parameter block, let $G_t \in \mathbb{R}^{m\times n}$ denote its stochastic gradient at iteration $t$.
Formally, Muon forms an orthogonalized update direction as
\begin{equation}
\label{eq:muon_direction_revised}
\begin{aligned}
M_t &= \beta M_{t-1} + G_t,\\
\widetilde{M}_t &= \beta M_t + G_t,\\
O_t &= \operatorname{msign}(\widetilde{M}_t)\in\mathbb{R}^{m\times n},
\end{aligned}
\end{equation}
where $\beta \in [0,1)$ is the momentum coefficient, and the operator $\operatorname{msign}(\cdot)$ returns the orthogonal polar factor of a matrix.
For a full-rank matrix $X$ with thin singular value decomposition (SVD) $X=U\Sigma V^\top$, this factor is
\begin{equation*}
    \operatorname{msign}(X)=UV^\top .
\end{equation*}
Equivalently, $\operatorname{msign}(\cdot)$ maps the nonzero singular values of $X$ to one, flattening the spectrum while retaining the singular vectors.
Rather than computing this SVD explicitly, Muon approximates this direction with the $K$-step Newton--Schulz operator:
\begin{equation}
\label{eq:ns_approx_msign}
    O_t
    =
    \operatorname{NS}_{K}(\widetilde{M}_t)
    \approx
    \operatorname{msign}(\widetilde{M}_t),
\end{equation}
where $\widetilde{M}_t$ is the momentum-based update proxy and $O_t$ is the Muon update direction.

\paragraph{Motivation for Variance-Adaptive Muon.}
The equal-$\beta$ reformulation in~\eqref{eq:variance-scaled} decomposes Adam's update into sign-based coordinate-wise normalization and variance-based noise-aware damping.
Muon extends the sign-normalization component to matrix-valued momentum via spectral normalization, but lacks variance estimates, leaving its update proxy without explicit variance-aware attenuation.
This gap motivates incorporating Adam-style variance modulation into Muon-style updates while preserving Muon's spectral normalization.

\section{Muon-NSR and Muon-VS}
\label{sec:method}

Motivated by the analysis, we introduce \textbf{Muon-NSR} and \textbf{Muon-VS}, two variance-adaptive extensions of Muon. 
Both methods apply variance-based modulation before Muon's spectral normalization step, preserving Muon's orthogonalized matrix-update structure while incorporating Adam-style adaptivity.
Algorithm~\ref{alg:muon_nsr} presents the procedure.

\begin{algorithm}[t]
\caption{Muon-NSR and Muon-VS.}
\label{alg:muon_nsr}
\begin{algorithmic}[1]
\REQUIRE Initial parameter $W_0\in\mathbb{R}^{m\times n}$; learning-rate schedule $\{\eta_t\}_{t\ge1}$; weight decay $\lambda_{\mathrm{wd}}$; shared EMA rate $\beta$; NSR coefficient $\gamma$ used only when $\tau=\mathrm{NSR}$; stabilizer $\varepsilon$; update scale $s_{\mathrm{scale}}$; Newton--Schulz operator $\operatorname{NS}_{K}(\cdot)$; variant selector $\tau\in\{\mathrm{NSR},\mathrm{VS}\}$.
\STATE $M_0 \leftarrow 0,\quad \Gamma_0 \leftarrow 0$
\FOR{$t=1,2,\ldots$}
    \STATE $G_t \leftarrow \nabla_{W}\mathcal{L}(W_{t-1})$
    \STATE $M_t \leftarrow \beta M_{t-1}+(1-\beta)G_t$
    \STATE $\Gamma_t \leftarrow \beta\Gamma_{t-1}+\beta(1-\beta)(M_{t-1}-G_t)^2$
    \STATE $\widehat{M}_t \leftarrow M_t/(1-\beta^t),\quad \widehat{\Gamma}_t \leftarrow \Gamma_t/(1-\beta^t)$
    \STATE $\widetilde{M}_t \leftarrow G_t+\frac{\beta}{1-\beta}\widehat{M}_t$
    \IF{$\tau=\mathrm{NSR}$}
    \STATE $M_{\tau,t} \leftarrow \frac{\widetilde{M}_t}{\sqrt{\widetilde{M}_t^2+\gamma\widehat{\Gamma}_t}+\varepsilon}$
    \ELSIF{$\tau=\mathrm{VS}$}
    \STATE $M_{\tau,t} \leftarrow \frac{\widetilde{M}_t}{\sqrt{\widehat{\Gamma}_t}+\varepsilon}$
\ENDIF
\STATE $O_t \leftarrow \operatorname{NS}_{K}(M_{\tau,t})$
\STATE $W_t \leftarrow W_{t-1}(1-\eta_t\lambda_{\mathrm{wd}})-\eta_t s_{\mathrm{scale}}O_t$
\ENDFOR
\end{algorithmic}
\end{algorithm}

\paragraph{Momentum and Variance.}
At iteration $t$, we maintain two coupled exponential moving statistics, following the derivation in Section~\ref{sec:prelim}.
The momentum estimator $M_t$ is updated according to~\eqref{eq:mk-update}, yielding a smoothed estimate of the local gradient mean under the shared EMA decay rate and serving as the directional signal for the subsequent polar-factor computation. 
In parallel, the variance surrogate $\Gamma_t$ is updated by~\eqref{eq:sig-update}. 
This recursion is driven by the element-wise squared deviation $(G_t-M_{t-1})^2$, which quantifies the discrepancy between the current stochastic gradient and the previous mean estimate. 
Consequently, $\Gamma_t$ estimates dispersion around the evolving mean, rather than tracking the raw second moment as in Adam. 
Together, $(M_t,\Gamma_t)$ provide the mean and variance information used for variance-adaptive normalization before applying the matrix-valued sign operator.

\paragraph{Nesterov-style acceleration.}
Following Muon's Nesterov update in~\eqref{eq:muon_direction_revised}, we form a lookahead proxy from the current gradient \(G_t\) and the EMA momentum \(M_t\).
Muon uses an unnormalized momentum recursion, whereas our formulation parameterizes the first-moment estimate as an Adam-style EMA.
Ignoring the bias correction in line~6 of Algorithm~\ref{alg:muon_nsr}, rescaling the EMA momentum in line~4 by \((1-\beta)^{-1}\) yields
\begin{equation*}
\frac{M_t}{1-\beta}
=
\beta \frac{M_{t-1}}{1-\beta}
+
G_t .
\end{equation*}
After this rescaling, the EMA momentum satisfies the same recursion as Muon's unnormalized momentum buffer.
The proxy in line~7 can therefore be expressed as
\begin{equation*}
\widetilde{M}_t
=
G_t
+
\beta \frac{M_t}{1-\beta},
\end{equation*}
which coincides with Muon's Nesterov-style proxy under the no-bias-correction reparameterization above.
Thus, the factor \((1-\beta)^{-1}\) reparameterizes the Adam-style EMA momentum into Muon's unnormalized momentum form, rather than acting as an update-scale heuristic.
The bias correction in line~6 compensates for the early-step underestimation induced by zero initialization.

\paragraph{Muon-NSR.}
Muon-NSR normalizes the Muon update proxy by incorporating the variance estimate $\widehat{\Gamma}_t$ into $\widetilde{M}_t$, following the variance-adaptive form in Eq.~\eqref{eq:variance-scaled}.
The normalized proxy is defined as
\begin{equation}
M_{\mathrm{NSR},t}
=
\frac{\widetilde{M}_t}
{\sqrt{\widetilde{M}_t^2+\gamma\widehat{\Gamma}_t}+\varepsilon},
\label{eq:method_nsr}
\end{equation}
where $\gamma\ge0$ controls the strength of variance-based damping and $\varepsilon>0$ ensures numerical stability.
Ignoring the stabilizer $\varepsilon$, each nonzero entry of $\widetilde{M}_t$ reduces to
\begin{equation*}
M_{\mathrm{NSR},t}
\approx
\frac{\operatorname{sign}(\widetilde{M}_t)}
{\sqrt{1+\gamma\widehat{\Gamma}_t/\widetilde{M}_t^2}},
\end{equation*}
which shows that Muon-NSR acts as a soft gate controlled by the estimated noise-to-signal ratio.
Entries with large estimated uncertainty relative to their signal strength are therefore damped in the pre-orthogonalization proxy.
Because this modulation is applied before Newton--Schulz orthogonalization, the adapted proxy is subsequently mapped into Muon's approximate polar-factor geometry.
Appendix~\ref{app:formal_ordering} provides a singular-value analysis of this ordering, while Section~\ref{subsec:order_NSR} tests it empirically.

\paragraph{Muon-VS.}
Muon-VS directly normalizes the update proxy using the estimated gradient variance:
\begin{equation*}
    M_{\mathrm{VS},t}
    =
    \frac{\widetilde{M}_t}
    {\sqrt{\widehat{\Gamma}_t}+\varepsilon}.
\end{equation*}
Compared with Muon-NSR, this formulation removes the squared proxy term $\widetilde{M}_t^2$ and the sensitivity coefficient $\gamma$ from the denominator.
This simplification reduces the optimizer-specific hyperparameter search space while preserving direct variance-based reweighting of the update proxy.
Ignoring the stabilizer $\varepsilon$, Muon-VS can be viewed as the large-$\gamma$ limiting case of Muon-NSR when $\gamma\widehat{\Gamma}_t \gg \widetilde{M}_t^2$, up to a positive scalar absorbed by the subsequent spectral normalization.

\begin{figure*}[t]
\centering
\includegraphics[width=\textwidth]{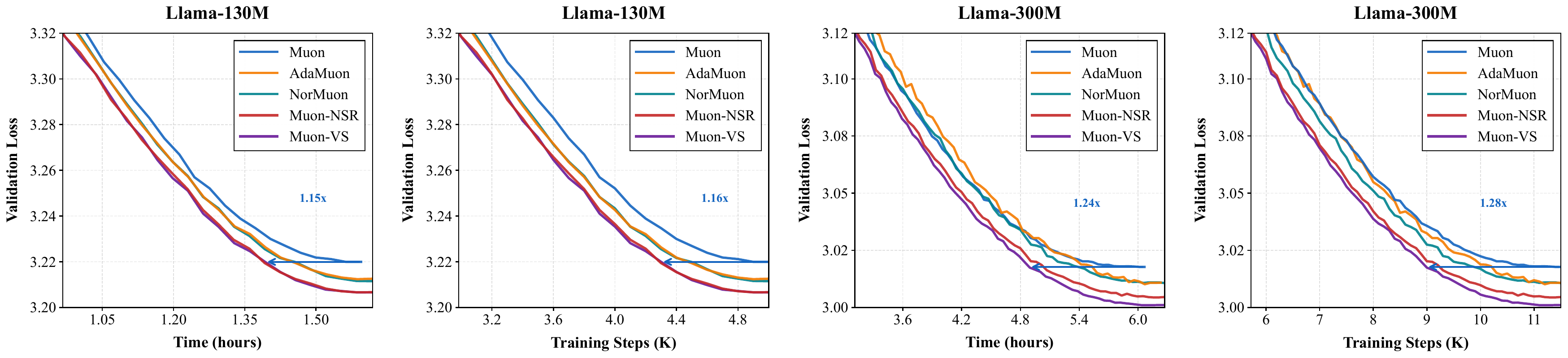}
\caption{\textbf{Suite~A Llama pretraining.}
Validation loss for \textbf{Llama-130M} and \textbf{Llama-300M} under the Suite~A protocol, plotted against wall-clock time and training steps.
The \textbf{Llama-1.2B} result is reported in Figure~\ref{fig:llm_stf_1.2B_val_loss_vs_iter}.}
\label{fig:suiteA_val_loss_time_steps}
\end{figure*}

\begin{figure*}[t]
\centering
\includegraphics[width=\textwidth]{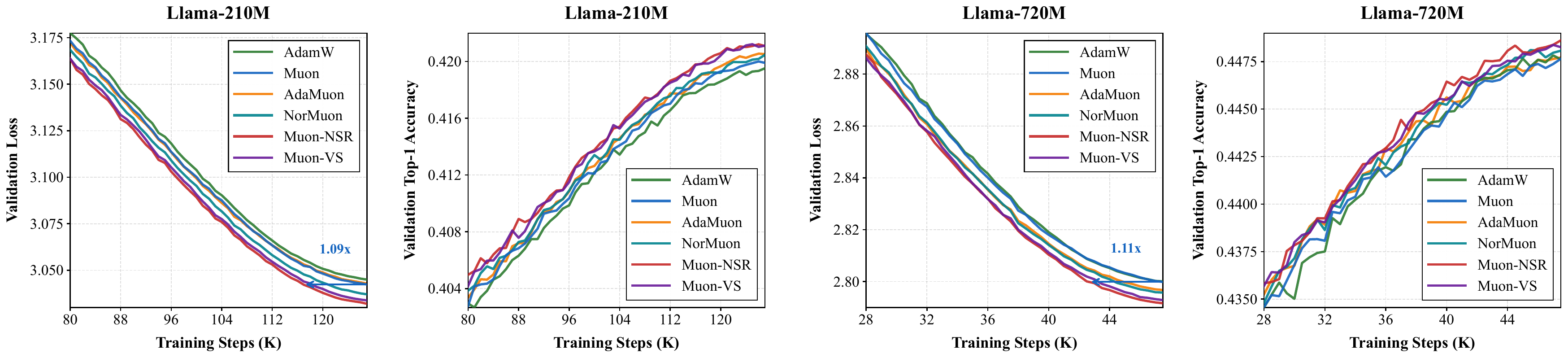}
\caption{\textbf{Suite~B Llama pretraining.}
Validation loss and validation-set next-token top-1 accuracy for \textbf{Llama-210M}
and \textbf{Llama-720M} under the Suite~B protocol.}
\label{fig:suiteB_val_loss_time_steps}
\end{figure*}

\begin{figure*}[t]
\centering
\includegraphics[width=\textwidth]{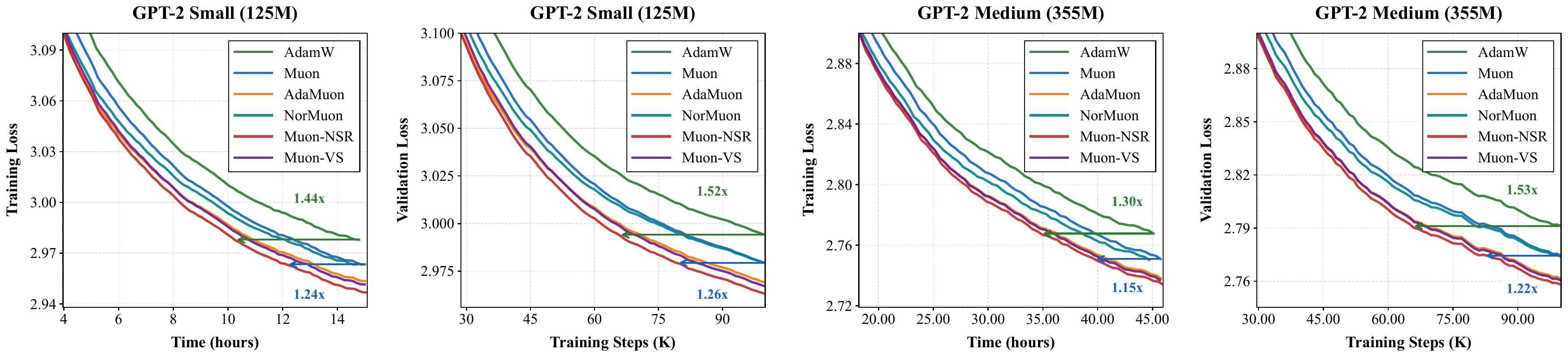}
\caption{\textbf{\texttt{nanoGPT}-based GPT-2 pretraining.}
Training loss versus wall-clock time and validation loss versus steps for \textbf{GPT-2 Small} and \textbf{GPT-2 Medium}, following the AdaMuon setup~\citep{si2025adamuon}, with EMA smoothing.}
\label{fig:nanogpt_loss_time_steps}
\end{figure*}

\paragraph{Orthogonalization and Parameter Update.}
\label{sec:orthogonalization_update}
Let $M_{\tau,t} \in \{M_{\mathrm{NSR},t}, M_{\mathrm{VS},t}\}$ denote the variance-modulated update proxy determined by the optimizer type $\tau$.
Following the Newton--Schulz approximation in~\eqref{eq:ns_approx_msign}, we compute the orthogonalized direction as
\begin{equation*}
    O_t = \operatorname{NS}_{K}(M_{\tau,t}) .
\end{equation*}
Let $W_t \in \mathbb{R}^{m\times n}$ denote the parameter matrix after the $t$-th update.
The parameter matrix is then updated using Muon's decoupled weight-decay rule:
\begin{equation*}
    W_t
    =
    W_{t-1}(1 - \eta_t \lambda_{\mathrm{wd}})
    - \eta_t \, s_{\mathrm{scale}}\, O_t,
\end{equation*}
where $\eta_t$ is the learning rate and $\lambda_{\mathrm{wd}} \ge 0$ is the weight-decay coefficient.
Since the norm of an orthogonalized matrix depends on its shape, practical Muon implementations usually apply a dimension-dependent scaling factor, such as
\begin{equation*}
    s_{\mathrm{scale}}
    \in
    \left\{
    0.2\sqrt{\max(m,n)},
    \sqrt{\max\left(1,\frac{m}{n}\right)}
    \right\}.
\end{equation*}
These rules compensate for shape-dependent update magnitudes while preserving the spectral normalization induced by $\operatorname{NS}_{K}(\cdot)$.
For a fair comparison, we adopt the same $s_{\mathrm{scale}}$ rule as the corresponding Muon baseline in our experiments, thereby more cleanly isolating the effect of pre-orthogonalization variance modulation.

\paragraph{Computational and Memory Cost.}
For an $m\times n$ parameter matrix, Muon-NSR and Muon-VS add an $O(mn)$ element-wise variance modulation step.
For typical matrix-shaped Transformer parameter groups, this added cost is dominated in practice by the $O(Kmn\min(m,n))$ cost of the $K$ Newton--Schulz iterations.
Both variants require one additional element-wise variance buffer and therefore have optimizer-state footprints comparable to those of Adam-type optimizers.
We report runtime and memory measurements in Appendix~\ref{app:optimizer_overhead}.

\section{Experiments}
\label{sec:exp}

We evaluate Muon-NSR and Muon-VS under fixed pretraining budgets across Llama-style and GPT-2 pretraining settings.
To assess optimization efficiency, we report fixed-budget final validation loss and step-to-target and time-to-target speedups using the reference baseline's final validation loss as the target.
Appendix~\ref{sec:supp_exp_details} provides the experimental settings, metric definitions, and implementation details, while Appendix~\ref{app:additional_details} presents additional results, including final validation losses and perplexities and, for the Llama-style settings, step-to-target and time-to-target speedups at five shared validation-loss targets spanning 20\%--100\% of Muon's training budget.
To assess whether the pretraining gains extend beyond validation loss, Appendix~\ref{app:small_model_downstream} provides a compact zero-shot downstream sanity check on the Suite~A Llama checkpoints.
The source code is publicly available at
\url{https://github.com/jingru-lee/Variance-Adaptive-Muon-EMNLP}.

\subsection{Llama Pretraining}
\label{subsec:exp_llama}

To assess whether the proposed methods generalize across Llama-style pretraining regimes, we conduct two complementary suites based on established training recipes.
Suite~A follows the pretraining optimizer protocol of \citet{wen2025fantastic}, while Suite~B follows the optimizer benchmarking setting of
\citet{semenov2025benchmarking}.
Both protocols include \emph{extensive hyperparameter tuning} and \emph{broad comparisons against more than ten optimizers}, providing rigorous and competitive evaluation settings.

\paragraph{Evaluation Scope.}
Rather than repeating a broad optimizer ranking, we build on the sweeps of \citet{wen2025fantastic} and \citet{semenov2025benchmarking}, which position Muon among the strongest pretraining optimizers under tuned protocols.
Accordingly, we evaluate whether variance-adaptive modulation improves Muon and compare our variants against Muon-family baselines under matched recipes.

\paragraph{Hyperparameter Reuse.}
Across all Llama experiments, Muon-NSR and Muon-VS inherit the corresponding protocol and Muon-tuned hyperparameters, with only one additional coarse coefficient \(\gamma\) for Muon-NSR and no optimizer-specific hyperparameter beyond Muon for Muon-VS.
This design evaluates whether the proposed variants can serve as drop-in alternatives to Muon-style optimizers with limited additional hyperparameter search.

\paragraph{Suite~A Setup.}
Following the Llama pretraining protocol of \citet{wen2025fantastic}, we evaluate Muon-NSR and Muon-VS under a 1$\times$ Chinchilla token budget on Llama-130M, Llama-300M, and Llama-1.2B using tuned Muon hyperparameters and the protocol's warmup--stable--decay (WSD) learning-rate schedule.
NorMuon is evaluated at all three scales using the Muon-tuned recipe, while AdaMuon is additionally included for Llama-130M and Llama-300M using the scaling rule selected by the learning-rate and scaling-rule search on Llama-130M in Appendix~\ref{sec:supp_suiteA_adamuon}, with a scale-adjusted learning rate for Llama-300M.
All runs use the Llama-2 tokenizer~\citep{touvron2023llama}, pretrain on DCLM~\citep{li2024datacomp}, and report validation loss on C4-EN~\citep{raffel2020exploring}.

\paragraph{Suite~A Results.}
Figure~\ref{fig:suiteA_val_loss_time_steps} shows the Suite~A pretraining validation-loss curves for Llama-130M and Llama-300M, while Figure~\ref{fig:llm_stf_1.2B_val_loss_vs_iter} reports the corresponding Llama-1.2B result.
Under the tuned Suite~A protocol~\citep{wen2025fantastic}, Muon-NSR and Muon-VS consistently improve over Muon at all three scales.
At Llama-130M and Llama-300M, where AdaMuon and NorMuon are also evaluated, both proposed variants achieve lower final validation loss than these adaptive Muon-family baselines, with corresponding gains also visible in the wall-clock curves.
At Llama-1.2B, Muon-VS outperforms NorMuon and achieves a
1.33$\times$ step-to-target speedup over the well-tuned Muon baseline, using the final Muon validation loss as the target.
Appendix~\ref{app:small_model_downstream} and Table~\ref{tab:small_model_zero_shot} further show consistent downstream macro-average gains over the evaluated Muon-family baselines on Suite~A Llama-130M and Llama-300M.
Appendix~\ref{app:muon2_supp} reports an additional comparison with Muon$^2$ on Suite~A Llama-130M and Llama-300M.

\paragraph{Suite~B Setup.}
Following the Llama pretraining benchmark protocol of \citet{semenov2025benchmarking}, we evaluate Llama-210M and Llama-720M on the FineWeb \texttt{sample-100BT} subset~\citep{penedo2024fineweb}.
The learning rate follows a linear-warmup cosine-decay schedule.
We compare Muon-NSR and Muon-VS against AdamW and representative Muon-family baselines under the matched benchmark recipe, with AdamW using the tuned benchmark hyperparameters and Muon-family methods using the same hyperparameter settings as Muon.

\paragraph{Suite~B Results.}
Figure~\ref{fig:suiteB_val_loss_time_steps} reports pretraining validation loss and validation-set next-token top-1 accuracy over training steps for Llama-210M and Llama-720M.
Under the matched Suite~B protocol, Muon-NSR and Muon-VS consistently rank among the leading methods at both model scales, outperforming AdamW, Muon, AdaMuon, and NorMuon.
Using the final Muon validation loss as the target, Muon-NSR achieves step-to-target speedups of \(1.09\times\) on Llama-210M and \(1.11\times\) on Llama-720M relative to the well-tuned Muon baseline.
The improvement is reflected in next-token top-1 accuracy across both settings, suggesting that the gains are not specific to the loss metric but reflect improvements in the optimization trajectory.

\subsection{GPT-2 Pretraining}
\label{subsec:exp_nanogpt}

\paragraph{GPT-2 Setup.}
We pretrain GPT-2 Small (125M) and Medium (355M) on OpenWebText~\citep{Gokaslan2019OpenWeb} using the AdaMuon~\citep{si2025adamuon} implementation built on \texttt{nanoGPT}~\citep{karpathy2022nanogpt}.
We compare Muon-NSR and Muon-VS against AdamW and representative Muon-family baselines under the same AdaMuon configuration, with a linear warmup followed by a constant learning rate throughout the remaining training steps.

\paragraph{GPT-2 Results.}
Figure~\ref{fig:nanogpt_loss_time_steps} reports training loss over wall-clock time and validation loss over training steps.
Across both GPT-2 Small and GPT-2 Medium, Muon-NSR achieves the fastest training-loss reduction and the lowest final validation loss among the evaluated optimizers.
Relative to the Muon baseline, Muon-NSR achieves time-to-target speedups of \(1.24\times\) and \(1.15\times\) on the training-loss curves and step-to-target speedups of \(1.26\times\) and \(1.22\times\) on the validation-loss curves for GPT-2 Small and GPT-2 Medium, respectively.
For reference, Figure~\ref{fig:nanogpt_loss_time_steps} also annotates the AdamW-relative speedups.
Muon-VS also attains lower final validation loss than AdamW, Muon, AdaMuon, and NorMuon at both scales.
Together, these results show that the benefits of variance-adaptive modulation extend to \texttt{nanoGPT}-based GPT-2 pretraining under the matched AdaMuon setup.
Appendix~\ref{app:muon2_supp} reports a comparison with Muon$^2$ on GPT-2.

\begin{table}[t]
\centering
\small
\setlength{\tabcolsep}{3.0pt}
\renewcommand{\arraystretch}{1.05}
\begin{tabular*}{\columnwidth}{
@{\extracolsep{\fill}}lcc@{}}
\toprule
Optimizer
& Val. Loss $\downarrow$
& $\Delta_{\mathrm{AdaMuon}}$ $\uparrow$ \\
\midrule
\multicolumn{3}{c}{\textbf{GPT-2 Small}} \\
AdaMuon (Baseline)
& 2.9572 & -- \\
Muon-NSR ($\gamma=1$)
& 2.9531 & +0.0041 \\
Muon-NSR ($\gamma=10$)
& 2.9508 & +0.0064 \\
Muon-NSR ($\gamma=100$)
& 2.9536 & +0.0036 \\
Muon-NSR ($\gamma=1000$)
& 2.9545 & +0.0027 \\
Muon-VS
& 2.9543 & +0.0029 \\
\midrule
\multicolumn{3}{c}{\textbf{Llama-130M}} \\
AdaMuon (Baseline)
& 3.2126 & -- \\
Muon-NSR ($\gamma=1$)
& 3.2104 & +0.0022 \\
Muon-NSR ($\gamma=10$)
& 3.2067 & +0.0059 \\
Muon-NSR ($\gamma=100$)
& 3.2072 & +0.0054 \\
Muon-NSR ($\gamma=1000$)
& 3.2067 & +0.0059 \\
Muon-NSR ($\gamma=10000$)
& 3.2077 & +0.0049 \\
Muon-VS
& 3.2066 & +0.0060 \\
\bottomrule
\end{tabular*}
\caption{\textbf{Sensitivity to the variance coefficient $\gamma$.} Final validation loss and absolute loss reduction $\Delta_{\mathrm{AdaMuon}}$ are reported relative to AdaMuon in both settings, with Muon-VS representing the large-$\gamma$ limiting behavior.}
\label{tab:optimizer_performance_comparison}
\end{table}

\subsection{Ablation Studies}
\label{sec:ablation_studies}

\paragraph{Role of the variance coefficient $\gamma$.}
\label{subsec:ablation_gamma}
Table~\ref{tab:optimizer_performance_comparison} examines the variance coefficient $\gamma$ in Eq.~\eqref{eq:method_nsr}. 
In Muon-NSR, $\gamma$ weights the variance surrogate $\widehat{\Gamma}_{t}$ in the denominator of Eq.~\eqref{eq:method_nsr}, thereby controlling the strength of variance-adaptive damping. 
We sweep $\gamma$ on a logarithmic grid while fixing all other settings to assess hyperparameter sensitivity.

The special case $\gamma=1$ corresponds to the most direct Adam--Muon combination under Adam's equal-beta variance decomposition, yielding the unit-coefficient form of the pre-orthogonalization normalization in Eq.~\eqref{eq:method_nsr}. Table~\ref{tab:optimizer_performance_comparison}
shows that this direct variant improves upon the AdaMuon baseline in both settings but remains suboptimal. Relaxing this unit coefficient yields lower validation loss on Llama-130M and GPT-2 Small. 
This gap motivates using $\gamma$ to control the strength of variance-adaptive damping, rather than fixing it to the value imposed by the combination.

Muon-NSR is robust to the choice of $\gamma$.
Across GPT-2 Small and Llama-130M, all evaluated values of $\gamma$ within the tested logarithmic ranges improve over AdaMuon.
Although the preferred setting can vary across training recipes, this variation does not indicate a need for fine-grained per-model tuning.
The performance variation across the tested values is mild, suggesting that a coarse log-scale choice is sufficient.
Thus, $\gamma$ mainly controls the strength of variance-adaptive damping rather than behaving as a brittle hyperparameter.
If eliminating this coefficient is desired, Muon-VS provides a $\gamma$-free alternative that removes the additional variance coefficient while retaining variance-based modulation.
Additional validation-loss trajectories for representative $\gamma$ settings are provided in Appendix~\ref{app:gamma_trajectory}.

\begin{figure}[t]
\centering
\includegraphics[width=\columnwidth]{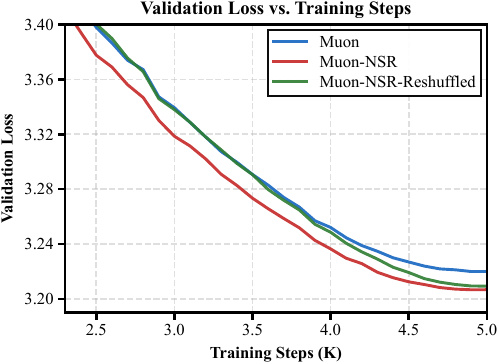}
\caption{\textbf{NSR ordering ablation on Suite~A Llama-130M.}
Validation loss trajectories of Muon, Muon-NSR, and the reshuffled post-orthogonalization variant.}
\label{fig:llm_stf_130M_val_loss_vs_iter_xiaorong}
\end{figure}

\paragraph{Ordering of NSR modulation.}
\label{subsec:order_NSR}
We examine whether variance modulation should be applied before or after Newton--Schulz orthogonalization.
Muon-NSR applies the NSR gate in Eq.~\eqref{eq:method_nsr} to the update proxy before the $K$-step Newton--Schulz map.
Under the exact polar map, all nonzero singular values are normalized to one, whereas a nonuniform element-wise gate applied afterward need not preserve this property.
Appendix~\ref{app:formal_ordering} establishes this structural distinction and derives a finite-step perturbation bound for the Newton--Schulz approximation used in practice.
For a controlled comparison, we consider a post-orthogonalization variant that first computes $O_t=\operatorname{NS}_{K}(\widetilde{M}_t)$ and then applies the stabilized coordinate-wise NSR gate as follows:
\begin{equation*}
S_t
=
\sqrt{
1 + \gamma
\frac{\widehat{\Gamma}_t}
{\widetilde{M}_t^{2}+\varepsilon}
},
\qquad
O^{\mathrm{post}}_t
=
O_t \oslash S_t,
\end{equation*}
where $\oslash$ denotes element-wise division.
Figure~\ref{fig:llm_stf_130M_val_loss_vs_iter_xiaorong}
shows that applying NSR modulation before orthogonalization yields a
smoother optimization trajectory and lower validation loss than applying
it afterward on Llama-130M. This comparison empirically supports
pre-orthogonalization variance adaptation as the preferred ordering in
this setting. Further algorithmic details and GPT-2 Small results are
provided in Appendix~\ref{sec:supp_algorithms}.

\section{Conclusion}

We presented Muon-NSR and Muon-VS, demonstrating that Adam-style variance adaptivity can be incorporated into Muon-style matrix updates while preserving Muon's orthogonalized update geometry.
Across Llama-style and GPT-2 pretraining settings at multiple model scales, both variants consistently improve convergence over a well-tuned Muon baseline and representative Muon-family baselines in the evaluated settings, achieving up to a 1.33$\times$ step-to-target speedup on Llama-1.2B, with Muon's final validation loss as the target.
These results position variance-adaptive modulation as an optimizer-level mechanism for improving the efficiency of Muon-based language model pretraining.

\section*{Limitations}

Due to computational constraints, our empirical
evaluation covers representative Llama-style and
GPT-2 pretraining settings up to 1.2B parameters.
These experiments provide controlled, matched comparisons across multiple model scales, but larger-scale settings such as 3B and 7B models remain beyond the scope of this study.
Evaluating Muon-NSR and Muon-VS in such regimes would require substantially larger training budgets and carefully matched tuning for all baselines.
We leave this larger-scale validation for future work.

\section*{Ethical Considerations}

This work studies optimizer design for efficient language model pretraining at scale.
The study does not involve human-subject data, create new annotations, or collect or release personally identifiable information.
We use publicly available pretraining corpora and validation sets
in accordance with the corresponding published protocols
solely for controlled aggregate model-level pretraining, ablation analysis,
and validation-loss evaluation.

The main anticipated benefit is a reduction in the number of optimization steps required to reach a fixed validation-loss target in the evaluated pretraining settings, thereby lowering computational cost under fixed budgets.
However, such efficiency gains may also lower the barrier to training large language models, making their broader impact application-dependent.
As this study uses publicly available web-scale corpora, it inherits the data-curation limitations of the original datasets.

We use all public datasets and codebases in accordance with their respective licenses and terms of use.
We do not redistribute the corpora or any pretrained checkpoints.
The released code includes third-party license and attribution information as well as data-preparation instructions for the datasets used in our experiments.

\section*{Acknowledgments}

H. Li was supported by the NSF China (No.~62476142).

\bibliography{references}

\clearpage

\appendix

\section{Experimental Setup Details}
\label{sec:supp_exp_details}

\paragraph{Overview and Compute Environment.}
This section details the Llama-style and GPT-2 pretraining settings considered in this study.
We describe the model architectures, data preprocessing, training budgets, batching strategies, optimizer hyperparameters, and evaluation protocols.
All pretraining experiments were implemented in PyTorch~\citep{paszke2019pytorch} and conducted on a single node with $8\times$ NVIDIA RTX~5090 GPUs using Distributed Data Parallel (DDP).

\subsection{Evaluation Metrics}
\label{app:evaluation_metrics}

For each run, we report the final validation loss under the fixed pretraining budget.
When reported, validation perplexity is computed as
$\mathrm{PPL}=\exp(\mathcal{L}_{\mathrm{val}})$,
where $\mathcal{L}_{\mathrm{val}}$ is the mean token-level cross-entropy on the validation set.

We quantify convergence efficiency using step-to-target and time-to-target speedups.
For a method $m$ and a baseline $b$, the speedup at target loss $\ell_\star$ is
\begin{equation*}
S_z(m,b;\ell_\star)
=
\frac{\tau_z(b;\ell_\star)}{\tau_z(m;\ell_\star)},
\qquad
z\in\{\mathrm{step},\mathrm{time}\},
\end{equation*}
where $\tau_z(a;\ell_\star)$ denotes the first interpolated step or wall-clock time at which method $a$ reaches $\ell_\star$ on the corresponding training- or validation-loss curve.
By construction, larger $S_z$ indicates faster convergence.
Unless otherwise specified, $\ell_\star$ is the reference baseline's final loss on the same curve under the same pretraining budget.
For the Llama-style multi-target analysis in Appendix~\ref{app:llama_multi_target_speedups}, we use Muon's
validation losses at 20\% intervals over 20\%--100\% of its
training budget as five shared targets.
For readability, the main-text speedups are annotated on the corresponding learning curves.

\begin{table*}[t]
\centering
\small
\setlength{\tabcolsep}{4pt}
\renewcommand{\arraystretch}{1.15}
\begin{tabular*}{\textwidth}{@{\extracolsep{\fill}} l *{12}{c} @{}}
\toprule
\textbf{Model} & \textbf{Hid} & \textbf{FFN} & \textbf{Heads} & \textbf{BSZ} & \textbf{$\eta_{\mathrm{adam}}$} & \textbf{$\eta_{\mathrm{muon}}$} & \textbf{$\beta_{\mathrm{muon}}$} & \textbf{$g_{\mathrm{norm}}$} & \textbf{$(\beta_1,\beta_2)$} & \textbf{WSD} & \textbf{WU} & \textbf{$\eta_{\min}$} \\
\midrule
Llama-130M & 512  & 2048 & 8  & 128 & 0.0032 & 0.016 & 0.95 & 1.0 & (0.8, 0.98) & 0.8 & 0 & 0 \\
Llama-300M & 768  & 3072 & 12 & 128 & 0.0024 & 0.008 & 0.98 & 1.0 & (0.8, 0.98) & 0.8 & 0 & 0 \\
Llama-1.2B & 1536 & 6144 & 24 & 256 & 0.0012 & 0.008 & 0.98 & 2.0 & (0.8, 0.98) & 1   & 0 & 0 \\
\bottomrule
\end{tabular*}
\caption{\textbf{Suite~A Llama pretraining configuration and optimizer hyperparameters.}
All models use 32 Transformer layers and a context length of $4096$.
Hid denotes the hidden size, FFN denotes the feed-forward network width, Heads denotes the number of attention heads, BSZ denotes the global batch size in sequences, $g_{\mathrm{norm}}$ denotes the global gradient-norm threshold, WU denotes warmup steps, and WSD denotes the decay-stage setting in the Suite~A protocol's warmup--stable--decay schedule, which uses linear decay.
}
\label{tab:suiteA_unified}
\end{table*}

\begin{table*}[t]
\centering
\footnotesize
\setlength{\tabcolsep}{4pt}
\renewcommand{\arraystretch}{1.15}
\begin{tabular*}{\textwidth}{@{\extracolsep{\fill}} l *{11}{c} @{}}
\toprule
\textbf{Model} & \textbf{Hid} & \textbf{L} & \textbf{H} & \textbf{Batch} & \textbf{Steps} & \textbf{Tok} & \textbf{Sch} & \textbf{WU} & \textbf{Peak} & \textbf{WD} & \textbf{Clip} \\
\midrule
\textbf{Llama-210M} & 768  & 24 & 12 & 256  & 128k & 16.8B & cos & 2k & $1\times 10^{-3}$ & 0.1 & 0.5 \\
\textbf{Llama-720M} & 2048 & 12 & 16 & 1984 & 48k  & 48.8B & cos & 2k & $1\times 10^{-3}$ & 0.1 & 0.1 \\
\bottomrule
\end{tabular*}
\caption{\textbf{Suite~B Llama pretraining configuration.}
Hid: hidden size; L: layers; H: attention heads;
Batch: global batch size in sequences; Tok: training tokens;
Sch: learning-rate decay schedule; WU: warmup steps;
Peak: peak learning rate; WD: weight decay;
Clip: gradient-norm clipping threshold.}
\label{tab:suiteB_llama_pretrain}
\end{table*}

\subsection{Suite~A Llama Pretraining Setup}
\label{sec:supp_suiteA_setup}


We conduct the Suite~A experiments with Llama architectures following the pretraining protocol of \citet{wen2025fantastic}, which prioritizes regime-specific hyperparameter optimization for fair benchmarking.
The codebase is built on the public pretraining scaffold released with \citet{liang2024cautious}.
The model-scale, training-budget, learning-rate-schedule, and Muon hyperparameter settings follow \citet{wen2025fantastic}, while the data and evaluation configuration is specified below.

\paragraph{Model Architecture.}
We pretrain three decoder-only models based on the Llama architecture: Llama-130M, Llama-300M, and Llama-1.2B.
To isolate the impact of width, we keep model depth and context length fixed across all scales.
Table~\ref{tab:suiteA_unified} provides the detailed architectural specifications.

\paragraph{Data and Preprocessing.}
We pretrain on DCLM using the Llama-2 tokenizer with sequence packing and
token-stream preprocessing.
Evaluation is performed on the C4-EN validation split.
Within each figure or table, all optimizers use the same preprocessing,
training budget, validation split, and training stream to ensure controlled
comparisons.

\paragraph{Optimizers and Hyperparameters.}
We evaluate Muon, NorMuon, Muon-NSR, and Muon-VS at all three
Suite~A scales, and include AdaMuon for Llama-130M and Llama-300M.
Muon-NSR, Muon-VS, and NorMuon reuse the Muon hyperparameters without separate retuning across scales, with $\gamma=1000$ for Muon-NSR in the main comparisons.
For AdaMuon, Appendix~\ref{sec:supp_suiteA_adamuon} reports a dedicated
learning-rate and scaling-rule search on Llama-130M, and the reported AdaMuon
results use the selected implementation-level scaling rule
\(s_{\mathrm{scale}} = \frac{0.2\sqrt{mn}}
{\left\|M_{\mathrm{update}}\right\|_{F}}\), where
\(M_{\mathrm{update}}\in\mathbb{R}^{m\times n}\) denotes AdaMuon's update matrix.
For Muon, Muon-NSR, Muon-VS, and NorMuon, we adopt the same Suite~A update-scale rule \(s_{\mathrm{scale}}=\sqrt{\max(1,m/n)}\) from \citet{wen2025fantastic}.

\subsection{Suite~B Llama Pretraining Setup}
\label{sec:supp_suiteB_setup}

We follow the Suite~B Llama pretraining benchmark~\citep{semenov2025benchmarking}.
All models use a decoder-only Llama architecture with Swish-gated linear unit (SwiGLU) feed-forward blocks, root-mean-square layer normalization (RMSNorm; $\varepsilon=10^{-5}$), Rotary Position Embeddings (RoPE)~\citep{su2024roformer}, tied input-output embeddings, bias-free linear layers, and no dropout.
Parameters are initialized from a truncated Gaussian distribution with $\sigma=0.02$.
The model configurations and training hyperparameters are summarized in Table~\ref{tab:suiteB_llama_pretrain}.

\paragraph{Data and Preprocessing.}
We pretrain on the FineWeb \texttt{sample-100BT} subset~\citep{penedo2024fineweb} following the benchmark preprocessing protocol.
The corpus is tokenized using the GPT-2 byte-pair encoding (BPE) tokenizer with a vocabulary size of $50{,}304$.
Both training and evaluation use a fixed context length of $512$ tokens.
The Llama-210M and Llama-720M runs use $16.8$B and $48.8$B training tokens, corresponding to $4\times$ and approximately $3.4\times$ their Chinchilla-optimal budgets of $4.2$B and $14.4$B tokens, respectively.

\paragraph{Optimization Protocol and Shared Settings.}
We compare six optimizers under a matched protocol: AdamW, Muon, AdaMuon, NorMuon, Muon-NSR, and Muon-VS.
The model architecture, data, training budget, and learning-rate schedule are held fixed across optimizers within each model configuration.
All optimizers use decoupled weight decay with $\lambda_{\mathrm{wd}}=0.1$ and model-specific global gradient-norm clipping thresholds, as reported in Table~\ref{tab:suiteB_llama_pretrain}.
The learning rate is linearly warmed up over $2{,}000$ steps to its peak value, followed by cosine decay to $0.01$ times the peak learning rate.
For Muon-family methods, we inherit the Muon baseline configuration except for method-specific components of the update rule.
Following the benchmark implementation~\citep{semenov2025benchmarking}, matrix-shaped parameters in Transformer blocks are optimized by the corresponding Muon-style optimizer, while embedding tables and vector-shaped parameters are updated by AdamW.
The Muon configuration uses momentum $\beta_{\mathrm{muon}}=0.95$, together with Nesterov acceleration and $K=5$ Newton--Schulz iterations.

\paragraph{Optimizer-Specific Settings.}
AdamW uses momentum coefficients $(\beta_1,\beta_2)=(0.9,0.999)$ for Llama-210M and $(0.9,0.99)$ for Llama-720M, with $\varepsilon_{\mathrm{adam}}=10^{-8}$.
For Muon, NorMuon, Muon-NSR, and Muon-VS, we use the benchmark Muon update-scaling rule $s_{\mathrm{scale}}=0.2\sqrt{\max(m,n)}$.
AdaMuon instead follows its original RMS-aligned update-scaling rule, $s_{\mathrm{scale}}=\frac{0.2\sqrt{mn}}{\left\|M_{\mathrm{update}}\right\|_{F}}$, as implemented in~\citet{si2025adamuon}.
For Muon-NSR, $\gamma$ is fixed to $1000$ at both model scales.

\paragraph{Evaluation Protocol.}
We evaluate all models under an identical validation protocol across optimizers.
We report validation loss, computed as the mean token-level cross-entropy, and next-token top-1 accuracy, computed as the fraction of token positions where the model's highest-probability prediction matches the ground-truth token.

\subsection{GPT-2 (\texttt{nanoGPT}) Pretraining Setup}
\label{sec:supp_nanogpt_setup}

We conduct GPT-2 pretraining experiments using the AdaMuon codebase~\citep{si2025adamuon}, which builds on the \texttt{nanoGPT} implementation~\citep{karpathy2022nanogpt} of GPT-2~\citep{radford2019language}.
Following AdaMuon, we use bias-free linear layers, disable dropout, retain Gaussian Error Linear Unit (GELU) activations, replace learned positional embeddings with RoPE, and replace the standard cosine decay schedule with a 2{,}000-step linear warmup followed by a constant learning rate.
We evaluate GPT-2 Small and GPT-2 Medium, with architectural configurations summarized in Table~\ref{tab:gpt2_model_configs}.

\paragraph{Data.}
We train on OpenWebText~\citep{Gokaslan2019OpenWeb} using the GPT-2 BPE tokenizer~\citep{radford2019language} and evaluate on its validation split.

\paragraph{Training Budget and Batching.}
All models are trained for 100{,}000 optimizer updates on 8 GPUs using Distributed Data Parallel (DDP).
We use gradient accumulation to obtain an effective batch size of 60 sequences per GPU, corresponding to an effective global batch size of 480 sequences per optimizer update.
With a sequence length of 1{,}024, this training budget contains about $4.92 \times 10^{10}$ tokens.

\paragraph{Optimizers and Hyperparameters.}
We compare our proposed Muon-NSR and Muon-VS with AdamW, Muon, AdaMuon, and NorMuon.
All methods use the same learning-rate schedule, with a peak learning rate of $6\times 10^{-4}$, weight decay of $0.1$, and a global gradient-norm clipping threshold of $1.0$.
AdamW uses $(\beta_1,\beta_2)=(0.9,0.95)$. 
For Muon-family methods, we follow the hybrid parameter partitioning strategy of \citet{si2025adamuon}: matrix-shaped parameters in Transformer blocks are optimized with Muon-style updates, while embedding tables and vector-shaped parameters are updated with AdamW.
We set the Muon momentum to $0.95$ and use $s_{\mathrm{scale}} = 0.2\sqrt{\max(m,n)}$, where $m$ and $n$ denote the matrix dimensions.
AdaMuon and NorMuon instead use the implementation-level scaling factor $s_{\mathrm{scale}} = \frac{0.2\sqrt{mn}}{\left\|M_{\mathrm{update}}\right\|_{F}}$.
For Muon-NSR, we set $\gamma=10$ in the main \texttt{nanoGPT} comparisons; ablation-specific values are specified separately.

\begin{table}[t]
\centering
\small
\setlength{\tabcolsep}{3pt}
\renewcommand{\arraystretch}{1.08}
\begin{tabular*}{\columnwidth}{@{\extracolsep{\fill}}lcccc@{}}
\toprule
\multicolumn{1}{c}{Model}
& \multicolumn{1}{c}{Layers}
& \multicolumn{1}{c}{Heads}
& \multicolumn{1}{c}{Hidden size}
& \multicolumn{1}{c}{Parameters} \\
\midrule
GPT-2 Small  & 12 & 12 & 768  & 125M \\
GPT-2 Medium & 24 & 16 & 1024 & 355M \\
\bottomrule
\end{tabular*}
\caption{\textbf{GPT-2 model configurations.}
Architecture sizes used in the \texttt{nanoGPT} pretraining experiments.}
\label{tab:gpt2_model_configs}
\end{table}

\begin{table}[t]
\centering
\small
\setlength{\tabcolsep}{3pt}
\renewcommand{\arraystretch}{1.12}
\begin{tabular*}{\columnwidth}{@{\extracolsep{\fill}}lccc@{}}
\toprule
Optimizer
& Val. Loss $\downarrow$
& Val. PPL $\downarrow$
& $\Delta_{\mathrm{Muon}}\,\uparrow$ \\
\midrule
\multicolumn{4}{c}{\textbf{Llama-130M}} \\
Muon     & 3.2199          & 25.03          & -- \\
AdaMuon  & 3.2126          & 24.84          & 0.0073 \\
NorMuon  & 3.2115          & 24.82          & 0.0084 \\
Muon-NSR & 3.2067          & 24.70          & 0.0132 \\
Muon-VS  & \textbf{3.2066} & \textbf{24.69} & \textbf{0.0133} \\
\midrule
\multicolumn{4}{c}{\textbf{Llama-300M}} \\
Muon     & 3.0177          & 20.44          & -- \\
AdaMuon  & 3.0109          & 20.31          & 0.0068 \\
NorMuon  & 3.0107          & 20.30          & 0.0070 \\
Muon-NSR & 3.0046          & 20.18          & 0.0131 \\
Muon-VS  & \textbf{3.0011} & \textbf{20.11} & \textbf{0.0166} \\
\midrule
\multicolumn{4}{c}{\textbf{Llama-1.2B}} \\
Muon     & 2.7458          & 15.58          & -- \\
NorMuon  & 2.7367          & 15.44          & 0.0091 \\
Muon-NSR & 2.7346          & 15.40          & 0.0112 \\
Muon-VS  & \textbf{2.7323} & \textbf{15.37} & \textbf{0.0135} \\
\bottomrule
\end{tabular*}
\caption{\textbf{Suite~A final validation loss and perplexity.}
Results are reported for each optimizer after the fixed pretraining budget at each model scale under the Suite~A protocol.
Perplexity is computed as $\mathrm{PPL}=\exp(\mathcal{L}_{\mathrm{val}})$.
$\Delta_{\mathrm{Muon}}=\mathcal{L}_{\mathrm{Muon}}-\mathcal{L}_{\mathrm{opt}}$ denotes the validation-loss reduction relative to Muon at the same model scale.
Bold indicates the best value within each scale.}
\label{tab:suiteA_final_loss}
\end{table}

\section{Additional Results and Comparisons}
\label{app:additional_details}

\subsection{Additional Suite~A Final-Loss Results}
\label{app:suiteA_final_loss}

\begin{figure}[t]
\centering
\includegraphics[width=\columnwidth]{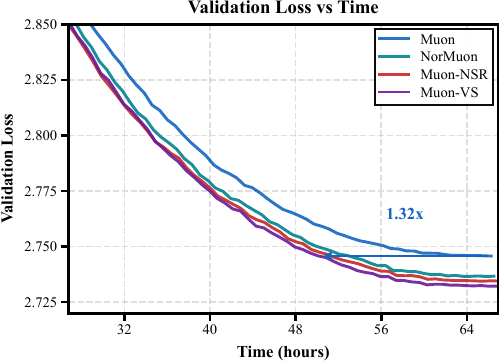}
\caption{\textbf{Suite~A Llama-1.2B wall-clock trajectory.}
Validation loss on the C4-EN validation split versus wall-clock time during pretraining on DCLM with the 1$\times$ Chinchilla token budget.}
\label{fig:llm_stf_1.2B_val_loss_vs_time}
\end{figure}

Table~\ref{tab:suiteA_final_loss} reports the final validation loss and perplexity under the fixed Suite~A pretraining budget.
Across all three model scales, Muon-NSR and Muon-VS improve over Muon; on Llama-130M and Llama-300M, they also outperform AdaMuon and NorMuon.
Together with the step-to-target and wall-clock results in Figures~\ref{fig:llm_stf_1.2B_val_loss_vs_iter}, \ref{fig:suiteA_val_loss_time_steps}, and~\ref{fig:llm_stf_1.2B_val_loss_vs_time}, these results show that pre-orthogonalization variance modulation consistently improves Muon-style optimization under the Suite~A protocol.

\begin{table}[t]
\centering
\small
\setlength{\tabcolsep}{2pt}
\renewcommand{\arraystretch}{1.05}
\begin{tabular*}{\columnwidth}{@{\extracolsep{\fill}}lccc@{}}
\toprule
Optimizer
& Val. Loss $\downarrow$
& Val. PPL $\downarrow$
& $\Delta_{\mathrm{AdamW}}\,\uparrow$ \\
\midrule
\multicolumn{4}{c}{\textbf{Llama-210M}} \\
AdamW    & 3.0450          & 21.01          & -- \\
Muon     & 3.0424          & 20.96          & 0.0026 \\
AdaMuon  & 3.0429          & 20.97          & 0.0021 \\
NorMuon  & 3.0371          & 20.84          & 0.0079 \\
Muon-NSR & \textbf{3.0320} & \textbf{20.74} & \textbf{0.0130} \\
Muon-VS  & 3.0338          & 20.78          & 0.0112 \\
\midrule
\multicolumn{4}{c}{\textbf{Llama-720M}} \\
AdamW    & 2.8001          & 16.45          & -- \\
Muon     & 2.7999          & 16.44          & 0.0002 \\
AdaMuon  & 2.7967          & 16.39          & 0.0034 \\
NorMuon  & 2.7957          & 16.37          & 0.0044 \\
Muon-NSR & \textbf{2.7916} & \textbf{16.31} & \textbf{0.0085} \\
Muon-VS  & 2.7929          & 16.33          & 0.0072 \\
\bottomrule
\end{tabular*}
\caption{\textbf{Suite~B final validation loss and perplexity.}
Results are reported for each optimizer after the fixed pretraining budget at each model scale under the Suite~B protocol.
Perplexity is computed as $\mathrm{PPL}=\exp(\mathcal{L}_{\mathrm{val}})$.
$\Delta_{\mathrm{AdamW}}=\mathcal{L}_{\mathrm{AdamW}}-\mathcal{L}_{\mathrm{opt}}$ denotes the validation-loss reduction relative to AdamW at the same model scale.
Bold indicates the best value within each scale.}
\label{tab:suiteB_final_loss}
\end{table}

\begin{table}[t]
\centering
\small
\setlength{\tabcolsep}{3pt}
\renewcommand{\arraystretch}{1.12}
\begin{tabular*}{\columnwidth}{@{\extracolsep{\fill}}lccc@{}}
\toprule
Optimizer
& Val. Loss $\downarrow$
& Val. PPL $\downarrow$
& $\Delta_{\mathrm{Muon}}\,\uparrow$ \\
\midrule
\multicolumn{4}{c}{\textbf{GPT-2 Small}} \\
AdamW    & 2.9810          & 19.71          & -0.0144 \\
Muon     & 2.9666          & 19.43          & -- \\
AdaMuon  & 2.9572          & 19.24          & 0.0094 \\
NorMuon  & 2.9671          & 19.44          & -0.0005 \\
Muon-NSR & \textbf{2.9508} & \textbf{19.12} & \textbf{0.0158} \\
Muon-VS  & 2.9543          & 19.19          & 0.0123 \\
\midrule
\multicolumn{4}{c}{\textbf{GPT-2 Medium}} \\
AdamW    & 2.7745          & 16.03          & -0.0187 \\
Muon     & 2.7558          & 15.73          & -- \\
AdaMuon  & 2.7429          & 15.53          & 0.0129 \\
NorMuon  & 2.7555          & 15.73          & 0.0003 \\
Muon-NSR & \textbf{2.7404} & \textbf{15.49} & \textbf{0.0154} \\
Muon-VS  & 2.7418          & 15.51          & 0.0140 \\
\bottomrule
\end{tabular*}
\caption{\textbf{GPT-2 final validation loss and perplexity.}
Results are reported for each optimizer after the fixed 100K-step pretraining budget at each model scale.
Perplexity is computed as $\mathrm{PPL}=\exp(\mathcal{L}_{\mathrm{val}})$.
$\Delta_{\mathrm{Muon}}=\mathcal{L}_{\mathrm{Muon}}-\mathcal{L}_{\mathrm{opt}}$ denotes the validation-loss reduction relative to Muon at the same model scale.
Bold indicates the best value within each scale.}
\label{tab:gpt2_final_loss}
\end{table}

\subsection{Additional Suite~B Final-Loss Results}
\label{app:suiteB_final_loss}

Table~\ref{tab:suiteB_final_loss} reports the Suite~B validation loss of AdamW, Muon, AdaMuon, NorMuon, and our proposed variants.
Figure~\ref{fig:suiteB_val_loss_time_steps} shows the corresponding learning curves, including AdaMuon and NorMuon.
Across both model scales, Muon yields only marginal gains over AdamW, whereas our variance-adaptive variants deliver clearer and larger reductions.
Muon-NSR attains the lowest final loss at both scales, while Muon-VS remains in the leading group and improves over all evaluated baselines.
These results indicate that incorporating variance statistics into Muon-style updates lowers final validation loss under a fixed training budget.

\begin{table*}[t]
\centering
\small
\setlength{\tabcolsep}{3.8pt}
\renewcommand{\arraystretch}{1.05}
\begin{tabular*}{\textwidth}{
@{\extracolsep{\fill}}llccccc@{}}
\toprule
Protocol / Model
& Quantity
& 20\%
& 40\%
& 60\%
& 80\%
& 100\% \\
\midrule

\multirow[c]{3}{*}{Suite~A Llama-130M}
& Target loss
& 3.7320 & 3.4783 & 3.3394 & 3.2521 & 3.2199 \\
& Step speedup
& \((1.07,\,1.07)\)
& \((1.07,\,1.07)\)
& \((1.05,\,1.06)\)
& \((1.05,\,1.06)\)
& \((1.17,\,1.16)\) \\
& Time speedup
& \((1.05,\,1.05)\)
& \((1.06,\,1.06)\)
& \((1.04,\,1.04)\)
& \((1.04,\,1.04)\)
& \((1.15,\,1.15)\) \\
\addlinespace[2pt]

\multirow[c]{3}{*}{Suite~A Llama-300M}
& Target loss
& 3.4120 & 3.1985 & 3.0929 & 3.0334 & 3.0177 \\
& Step speedup
& \((1.10,\,1.09)\)
& \((1.07,\,1.08)\)
& \((1.08,\,1.09)\)
& \((1.10,\,1.11)\)
& \((1.24,\,1.28)\) \\
& Time speedup
& \((1.06,\,1.05)\)
& \((1.03,\,1.04)\)
& \((1.04,\,1.06)\)
& \((1.06,\,1.07)\)
& \((1.20,\,1.24)\) \\
\addlinespace[2pt]

\multirow[c]{3}{*}{Suite~A Llama-1.2B}
& Target loss
& 3.0167 & 2.8666 & 2.7901 & 2.7542 & 2.7458 \\
& Step speedup
& \((1.08,\,1.07)\)
& \((1.08,\,1.07)\)
& \((1.09,\,1.10)\)
& \((1.14,\,1.15)\)
& \((1.30,\,1.33)\) \\
& Time speedup
& \((1.11,\,1.10)\)
& \((1.08,\,1.07)\)
& \((1.09,\,1.10)\)
& \((1.13,\,1.15)\)
& \((1.29,\,1.32)\) \\
\midrule

\multirow[c]{3}{*}{Suite~B Llama-210M}
& Target loss
& 3.3593 & 3.2695 & 3.1849 & 3.0922 & 3.0424 \\
& Step speedup
& \((1.14,\,1.14)\)
& \((1.10,\,1.10)\)
& \((1.04,\,1.04)\)
& \((1.03,\,1.03)\)
& \((1.09,\,1.08)\) \\
& Time speedup
& \((1.14,\,1.15)\)
& \((1.11,\,1.10)\)
& \((1.03,\,1.03)\)
& \((1.03,\,1.03)\)
& \((1.08,\,1.07)\) \\
\addlinespace[2pt]

\multirow[c]{3}{*}{Suite~B Llama-720M}
& Target loss
& 3.0582 & 2.9608 & 2.8896 & 2.8261 & 2.7999 \\
& Step speedup
& \((1.06,\,1.06)\)
& \((1.07,\,1.08)\)
& \((1.04,\,1.04)\)
& \((1.04,\,1.04)\)
& \((1.11,\,1.10)\) \\
& Time speedup
& \((1.10,\,1.10)\)
& \((1.09,\,1.09)\)
& \((1.04,\,1.05)\)
& \((1.04,\,1.04)\)
& \((1.12,\,1.10)\) \\
\bottomrule
\end{tabular*}
\caption{\textbf{Step-to-target and time-to-target speedups across
shared validation-loss targets for Llama-style pretraining.}
Targets are selected from Muon's validation losses at 20\%
increments from 20\% to 100\% of its training budget.
Each speedup entry is ordered as (Muon-NSR, Muon-VS) relative
to Muon (\(1.00\)).}
\label{tab:llama_multi_target_speedups}
\end{table*}

\subsection{Multi-Target Speedups on Llama-Style Pretraining}
\label{app:llama_multi_target_speedups}

Because step-to-target and time-to-target speedups depend on the validation-loss target, we complement the single-target speedups reported in the main figures with speedups at five shared validation-loss targets to assess target robustness across all Llama-style settings in Suites~A and~B.
For each setting, we define the five targets as Muon's validation losses measured at 20\%, 40\%, 60\%, 80\%, and 100\% of its training budget, and use the same targets to evaluate Muon, Muon-NSR, and Muon-VS.
The first step and wall-clock time at which each method reaches a target are estimated by linear interpolation between adjacent evaluations, following Appendix~\ref{app:evaluation_metrics}.

Table~\ref{tab:llama_multi_target_speedups} shows that both proposed variants consistently require fewer training steps and less wall-clock time than Muon to reach every reported target across all evaluated settings in both pretraining suites.
However, the magnitude of the speedup varies across target losses and does not necessarily change monotonically as the target is moved later in training.
For example, on Suite~B Llama-210M, the step-speedup tuple decreases from \((1.14\times, 1.14\times)\) at the 20\% target to \((1.09\times, 1.08\times)\) at the 100\% target.
Taken together with the fixed-budget final-loss results in Tables~\ref{tab:suiteA_final_loss} and~\ref{tab:suiteB_final_loss}, the multi-target analysis shows that the observed acceleration is not specific to a single endpoint target.
Both variants reach matched validation-loss levels earlier across all five evaluated targets while also attaining lower final validation loss.

\subsection{Additional GPT-2 Final-Loss Results}
\label{app:gpt2_final_loss}

Table~\ref{tab:gpt2_final_loss} reports the final validation loss and perplexity for GPT-2 Small and GPT-2 Medium under the fixed 100K-step training budget.
The values are taken from the final unsmoothed validation log point for each optimizer run, complementing the smoothed training and validation curves in Figure~\ref{fig:nanogpt_loss_time_steps}.
At both model scales, Muon-NSR achieves the lowest validation loss, while Muon-VS also improves over AdamW, Muon, AdaMuon, and NorMuon.
These final-loss results confirm that the GPT-2 gains observed in the learning curves persist at the end of training.

\begin{table}[t]
\centering
\small
\setlength{\tabcolsep}{3pt}
\renewcommand{\arraystretch}{1.18}
\begin{tabular*}{\columnwidth}{@{\extracolsep{\fill}}lcc@{}}
\toprule
\textbf{Learning Rate}
& \textbf{Val. Loss} $\downarrow$
& \textbf{Val. PPL} $\downarrow$ \\
\midrule
\multicolumn{3}{c}{\textbf{Scale I: }$s_{\mathrm{scale}} = \frac{0.2\sqrt{mn}}{\left\|M_{\mathrm{update}}\right\|_{F}}$} \\
$1.0\times10^{-3}$          & 3.2486          & 25.75 \\
\textbf{$2.0\times10^{-3}$} & \textbf{3.2126} & \textbf{24.84} \\
$3.0\times10^{-3}$          & 3.2143          & 24.89 \\
$4.0\times10^{-3}$          & 3.2151          & 24.91 \\
\midrule
\multicolumn{3}{c}{\textbf{Scale II: }$s_{\mathrm{scale}}=\sqrt{\max(1,m/n)}$} \\
$2.0\times10^{-4}$ & 3.2441 & 25.64 \\
$3.0\times10^{-4}$ & 3.2266 & 25.19 \\
$4.0\times10^{-4}$ & 3.2230 & 25.10 \\
$5.0\times10^{-4}$ & 3.2258 & 25.17 \\
\bottomrule
\end{tabular*}
\caption{\textbf{AdaMuon learning-rate and scaling-rule search on Suite~A Llama-130M.}
Validation loss and perplexity are reported for each configuration.
Perplexity is $\mathrm{PPL}=\exp(\mathcal{L}_{\mathrm{val}})$.
The best configuration used in the Suite~A comparison is highlighted in bold.}
\label{tab:adamuon_grid_search}
\end{table}

\subsection{AdaMuon Tuning Details on Suite~A}
\label{sec:supp_suiteA_adamuon}

\paragraph{Motivation.}
The Suite~A experiments include AdaMuon as an adaptive Muon-family baseline for Llama-130M and Llama-300M.
Directly reusing the Suite~A Muon learning rate and update-scale rule is suboptimal for AdaMuon on Llama-130M.
We therefore perform a dedicated search over the AdaMuon learning rate and update-scale rule for matrix-shaped parameters on Llama-130M, use the selected configuration at that scale, and transfer the selected scaling rule with a scale-adjusted learning rate to Llama-300M.
We also attempted limited AdaMuon trials at Llama-1.2B, but the available compute budget did not permit a broad optimizer-specific search at this larger model scale.

\begin{table}[t]
\centering
\small
\setlength{\tabcolsep}{3pt}
\renewcommand{\arraystretch}{1.12}
\begin{tabular*}{\columnwidth}{@{\extracolsep{\fill}}lccc@{}}
\toprule
\textbf{Optimizer}
& \textbf{Main Run} $\downarrow$
& \textbf{Repeated Run} $\downarrow$
& $\boldsymbol{\Delta_{\mathrm{repeat}}}$ \\
\midrule
\multicolumn{4}{c}{\textbf{Suite~B Llama-210M}} \\
Muon-NSR & 3.0320 & 3.0328 & 0.0008 \\
Muon-VS  & 3.0338 & 3.0328 & 0.0010 \\
\midrule
\multicolumn{4}{c}{\textbf{Suite~B Llama-720M}} \\
Muon-NSR & 2.7916 & 2.7945 & 0.0029 \\
Muon-VS  & 2.7929 & 2.7927 & 0.0002 \\
\midrule
\multicolumn{4}{c}{\textbf{GPT-2 Small}} \\
Muon-NSR & 2.9508 & 2.9502 & 0.0006 \\
Muon-VS  & 2.9543 & 2.9550 & 0.0007 \\
\midrule
\multicolumn{4}{c}{\textbf{GPT-2 Medium}} \\
Muon-NSR & 2.7404 & 2.7393 & 0.0011 \\
Muon-VS  & 2.7418 & 2.7417 & 0.0001 \\
\bottomrule
\end{tabular*}
\caption{\textbf{Repeat-run consistency check.}
We repeat Muon-NSR and Muon-VS on Suite~B and GPT-2 using the main-experiment recipe, hyperparameters, and nominal random seed.
We report main- and repeat-run validation losses; $\Delta_{\mathrm{repeat}}$ is their absolute difference.}
\label{tab:repeat_run_validation}
\end{table}


\paragraph{AdaMuon search protocol.}
Table~\ref{tab:adamuon_grid_search} reports two learning-rate sweeps under two update-scale rules: the AdaMuon implementation-level rule, \(s_{\mathrm{scale}} = \frac{0.2\sqrt{mn}}{\left\|M_{\mathrm{update}}\right\|_{F}}\), and the Suite~A Muon rule, \(s_{\mathrm{scale}}=\sqrt{\max(1,m/n)}\).
Because these rules induce different update magnitudes, we use rule-specific learning-rate ranges, with all other settings fixed under the Suite~A protocol.
The best overall configuration uses the AdaMuon implementation-level scaling rule with a learning rate of \(2.0\times10^{-3}\), which we adopt for Llama-130M.
For Llama-300M, we retain the same scaling rule and set the AdaMuon learning rate to \(1.3\times10^{-3}\) based on the Llama-130M tuning result and model-scale extrapolation.

\begin{figure*}[t]
\centering
\includegraphics[width=\textwidth]{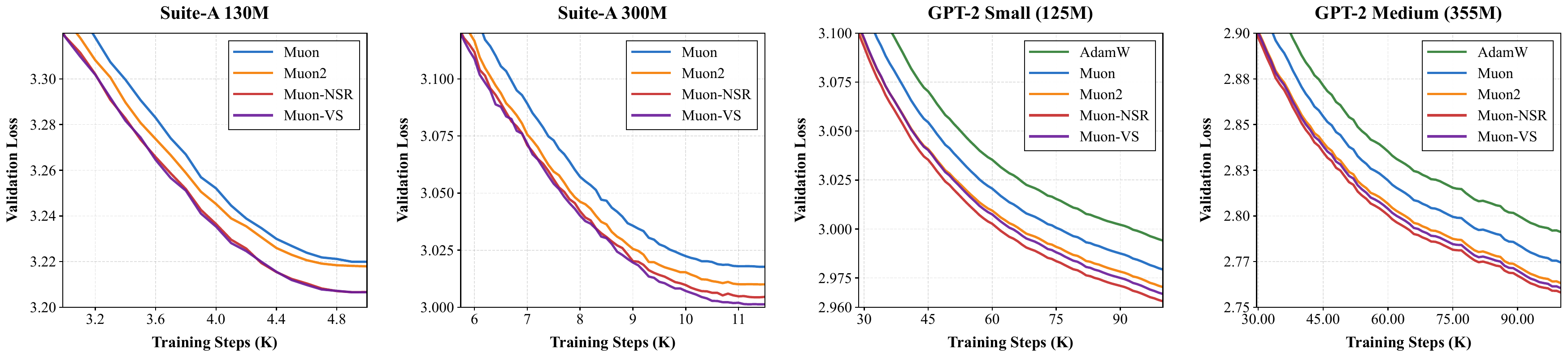}
\caption{\textbf{Supplementary Muon$^2$ comparison on Suite~A and GPT-2.}
Validation loss over training steps for Suite~A Llama-130M, Suite~A Llama-300M, GPT-2 Small (125M), and GPT-2 Medium (355M).
All panels compare Muon, Muon$^2$, Muon-NSR, and Muon-VS; the GPT-2 panels additionally include AdamW.
Lower is better.}
\label{fig:muon2_gpt_suitea}
\end{figure*}

\subsection{\texorpdfstring{Supplementary Comparison with Muon$^2$ on GPT-2 and Suite~A}{Supplementary Comparison with Muon2 on GPT-2 and Suite~A}}
\label{app:muon2_supp}


\paragraph{Motivation.}
Muon$^2$ is a recent Muon-family optimizer that preconditions the momentum proxy with an Adam-style uncentered second-moment estimate before Newton--Schulz orthogonalization, making it a relevant pre-orthogonalization adaptive baseline.
By contrast, Muon$^2$ targets better conditioning of this input, whereas Muon-NSR and Muon-VS are motivated by Adam's variance-adaptive sign-update interpretation and use online centered variance estimates to attenuate noise-dominated entries before orthogonalization.

\paragraph{Setup.}
We compare validation loss over training steps across four representative settings: Suite~A Llama-130M, Suite~A Llama-300M, GPT-2 Small, and GPT-2 Medium.
For each setting, Muon$^2$ follows the same model, data, training budget, learning-rate schedule, and parameter-partitioning scheme as the corresponding Muon-family setting.
This comparison is intended as a matched-protocol evaluation rather than a fully retuned optimizer study.
We also evaluated Muon$^2$ on Suite~A Llama-1.2B using the corresponding Muon configuration.
Because its trajectory closely overlaps with Muon and no optimizer-specific tuning was conducted at this scale, we omit this curve from the figure.

\paragraph{Findings.}
Figure~\ref{fig:muon2_gpt_suitea} shows that Muon$^2$ improves over Muon across all four supplementary settings, confirming it as a competitive adaptive baseline before orthogonalization.
Under the same matched protocols, Muon-NSR and Muon-VS achieve lower final validation loss than Muon$^2$ across all four settings.
This trend is observed across both the Suite~A Llama and GPT-2 experiments, despite their different model families and training recipes.
This supplementary comparison therefore extends the matched-protocol evaluation to a recent adaptive Muon-family baseline that also operates before orthogonalization and provides additional evidence for the effectiveness of the proposed variants under matched training conditions.

\subsection{Repeat-Run Consistency and Multi-Seed Robustness}
\label{app:repeat_run}

\paragraph{Same-nominal-seed repeat runs.}
Because repeating all optimizer--scale combinations across multiple independent seeds is computationally impractical, we use same-nominal-seed reruns as a repeatability check rather than an independent-seed robustness analysis.
We perform one additional end-to-end run for each of Muon-NSR and Muon-VS on four settings with relatively small performance margins: Suite~B Llama-210M, Suite~B Llama-720M, GPT-2 Small, and GPT-2 Medium.
Each repeated run uses the same data, model configuration, training budget, learning-rate schedule, parameter partitioning, optimizer hyperparameters, and nominal seed as the corresponding main experiment.
As shown in Table~\ref{tab:repeat_run_validation}, all repeated runs remain close to their main-run counterparts, with $\Delta_{\mathrm{repeat}} \leq 0.0029$.
They also remain below the corresponding non-proposed baselines reported in Tables~\ref{tab:suiteB_final_loss} and~\ref{tab:gpt2_final_loss}.
These small deviations support the repeatability of the reported results under the same nominal seed and do not alter the conclusions.

\begin{figure}[t]
\centering
\includegraphics[width=\columnwidth]
{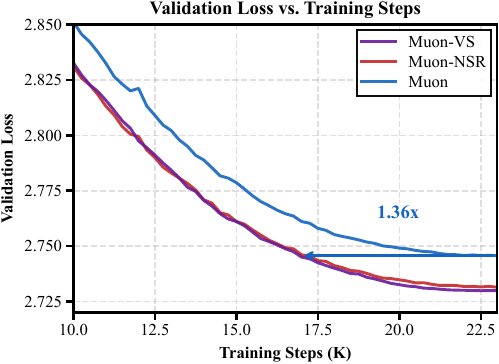}
\caption{\textbf{Data-subset robustness on Suite~A Llama-1.2B.}
Validation loss over training steps on a different fixed DCLM
subset, with all other settings matched to the primary
Suite~A Llama-1.2B comparison.}
\label{fig:llm_stf_1.2B_val_loss_vs_iter_robustness}
\end{figure}

\paragraph{Data-subset robustness on Suite~A Llama-1.2B.}
Because the DCLM corpus substantially exceeds the $1\times$ Chinchilla training budget used in Suite~A, we additionally assess sensitivity to the sampled training data by evaluating Muon, Muon-NSR, and Muon-VS on an alternative fixed DCLM subset, with all other experimental settings held fixed.
As shown in Figure~\ref{fig:llm_stf_1.2B_val_loss_vs_iter_robustness}, Muon-VS achieves a $1.36\times$ step-to-target speedup over Muon on this alternative subset, compared with $1.33\times$ in the primary four-optimizer comparison in Figure~\ref{fig:llm_stf_1.2B_val_loss_vs_iter}.
The comparable gains across the two fixed subsets suggest that the observed acceleration is not driven by a particular sampled DCLM subset.

\paragraph{Eight-seed robustness on Suite~A Llama-130M.}
For computational tractability, we conduct the eight-seed evaluation on Suite~A Llama-130M, comparing Muon, AdaMuon, NorMuon, Muon$^2$, Muon-NSR, and Muon-VS under a matched protocol.
All optimizers use the same eight random seeds, with the model, data, training budget, scheduler, parameter partitioning, and evaluation settings matched across optimizers and optimizer-specific hyperparameters fixed across seeds.
At each checkpoint, we report mean validation loss with its pointwise two-sided 95\% Student-$t$ confidence interval, $\bar{L} \pm t_{0.975,\,7}\,\frac{s}{\sqrt{8}}$, where $s$ is the sample standard deviation across seeds.
As shown in Table~\ref{tab:multiseed_suita_130m}, Muon-NSR and Muon-VS achieve the two lowest mean validation losses at all five reported checkpoints (1K--5K steps), supporting the robustness of the improvements across random seeds.

\begin{table*}[t]
\centering
\small
\setlength{\tabcolsep}{3.8pt}
\renewcommand{\arraystretch}{1.08}
\begin{tabular*}{\textwidth}{@{\extracolsep{\fill}}lccccc@{}}
\toprule
Optimizer
& 20\% (1K)
& 40\% (2K)
& 60\% (3K)
& 80\% (4K)
& 100\% (5K) \\
\midrule
Muon
& $3.7292 \pm 0.0098$
& $3.4733 \pm 0.0049$
& $3.3376 \pm 0.0022$
& $3.2488 \pm 0.0026$
& $3.2182 \pm 0.0028$ \\

AdaMuon
& $3.7418 \pm 0.0047$
& $3.4662 \pm 0.0026$
& $3.3276 \pm 0.0020$
& $3.2405 \pm 0.0021$
& $3.2115 \pm 0.0020$ \\

NorMuon
& $3.7090 \pm 0.0062$
& $3.4603 \pm 0.0024$
& $3.3271 \pm 0.0027$
& $3.2395 \pm 0.0031$
& $3.2095 \pm 0.0035$ \\

Muon$^2$
& $3.7119 \pm 0.0075$
& $3.4573 \pm 0.0044$
& $3.3234 \pm 0.0029$
& $3.2393 \pm 0.0039$
& $3.2126 \pm 0.0040$ \\

Muon-NSR
& $3.7015 \pm 0.0060$
& $\mathbf{3.4498 \pm 0.0031}$
& $\mathbf{3.3177 \pm 0.0027}$
& $\mathbf{3.2324 \pm 0.0031}$
& $\mathbf{3.2036 \pm 0.0024}$ \\

Muon-VS
& $\mathbf{3.7001 \pm 0.0070}$
& $3.4517 \pm 0.0027$
& $3.3194 \pm 0.0037$
& $3.2330 \pm 0.0034$
& $3.2037 \pm 0.0032$ \\
\bottomrule
\end{tabular*}
\caption{\textbf{Eight-seed validation-loss results on Suite~A Llama-130M.}
Entries report mean validation loss over the same eight random seeds across optimizers $\pm$ the half-width of the pointwise two-sided 95\% Student-$t$ confidence interval.
Bold marks the lowest mean validation loss at each checkpoint.}
\label{tab:multiseed_suita_130m}
\end{table*}

\begin{figure}[t]
\centering
\includegraphics[width=\columnwidth]{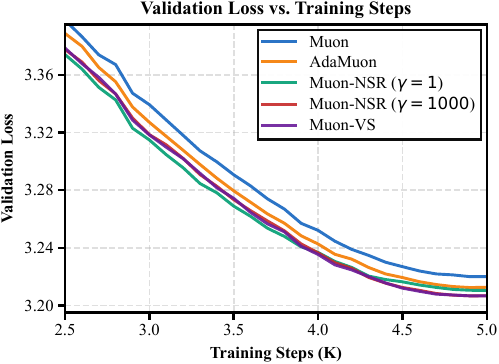}
\caption{\textbf{Representative $\gamma$ trajectories on Suite~A Llama-130M.}
Validation loss for Muon, AdaMuon, Muon-NSR
($\gamma\in\{1,1000\}$), and Muon-VS.}
\label{fig:appendix_gamma_trajectory}
\end{figure}

\section{Analysis and Ablation Studies}

\subsection{Validation-Loss Trajectories for \texorpdfstring{$\gamma$}{gamma}}
\label{app:gamma_trajectory}

To complement the final-loss comparison in Table~\ref{tab:optimizer_performance_comparison}, Figure~\ref{fig:appendix_gamma_trajectory} shows validation-loss trajectories for representative $\gamma$ settings on Suite~A Llama-130M.
The direct Adam--Muon variant, Muon-NSR ($\gamma=1$), achieves lower final validation loss than both Muon and AdaMuon, but remains above Muon-NSR ($\gamma=1000$) and Muon-VS.
This result suggests that direct Adam-style variance normalization is beneficial, but does not recover the best performance observed among the evaluated variance-adaptive settings.
These trajectories are consistent with the sensitivity results in Table~\ref{tab:optimizer_performance_comparison}.

\begin{algorithm}[t]
\caption{Muon-NSR-Reshuffled.}
\label{alg:muon_nsr_reshuffled}
\begin{algorithmic}[1]
\REQUIRE Initial parameter $W_0\in\mathbb{R}^{m\times n}$; learning-rate schedule $\{\eta_t\}_{t\ge1}$; weight decay $\lambda_{\mathrm{wd}}$; shared EMA rate $\beta$; NSR coefficient $\gamma$; stabilizer $\varepsilon$; update scale $s_{\mathrm{scale}}$; Newton--Schulz operator $\operatorname{NS}_{K}(\cdot)$.
\STATE $M_0 \leftarrow 0,\quad \Gamma_0 \leftarrow 0$
\FOR{$t=1,2,\ldots$}
    \STATE $G_t \leftarrow \nabla_{W}\mathcal{L}(W_{t-1})$
    \STATE $M_t \leftarrow \beta M_{t-1}+(1-\beta)G_t$
    \STATE $\Gamma_t \leftarrow \beta\Gamma_{t-1}+\beta(1-\beta)(M_{t-1}-G_t)^2$
    \STATE $\widehat{M}_t \leftarrow M_t/(1-\beta^t),\quad \widehat{\Gamma}_t \leftarrow \Gamma_t/(1-\beta^t)$
    \STATE $\widetilde{M}_t \leftarrow G_t+\frac{\beta}{1-\beta}\widehat{M}_t$
    \STATE $O_t \leftarrow \operatorname{NS}_{K}(\widetilde{M}_t)$
    \STATE $S_t \leftarrow \sqrt{1+\gamma\,\dfrac{\widehat{\Gamma}_t}{\widetilde{M}_t^2+\varepsilon}}$ \COMMENT{element-wise}
    \STATE $O_t^{\mathrm{post}} \leftarrow O_t \oslash S_t$ \COMMENT{element-wise}
    \STATE $W_t \leftarrow W_{t-1}(1-\eta_t\lambda_{\mathrm{wd}})-\eta_t s_{\mathrm{scale}}O_t^{\mathrm{post}}$
\ENDFOR
\end{algorithmic}
\end{algorithm}

\begin{figure}[t]
\centering
\includegraphics[width=\columnwidth]{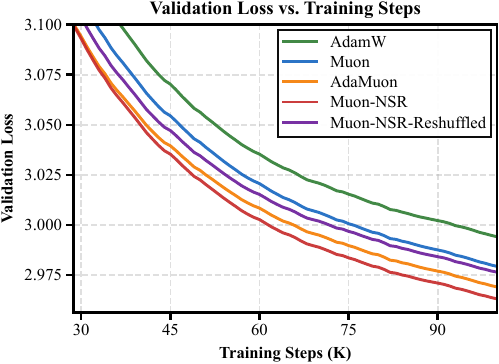}
\caption{\textbf{NSR ordering ablation on GPT-2 Small.}
Validation-loss trajectories for Muon-NSR and its post-orthogonalization variant, Muon-NSR-Reshuffled, with AdamW, Muon, and AdaMuon as baselines.}
\label{fig:gpt2_125m_reshuffled}
\end{figure}

\subsection{Pre- and Post-Orthogonalization Variance Modulation}
\label{app:formal_ordering}

We formalize the distinction between pre- and post-orthogonalization variance modulation in terms of singular-value structure.
Applying an element-wise gate before the exact polar map yields a polar update with equal nonzero singular values, whereas a nonuniform gate applied afterward need not preserve this property.
We derive a perturbation bound relating the Newton--Schulz approximation error to deviations from this singular-value structure in the practical finite-step setting.

Let $Z\in\mathbb{R}^{m\times n}$ have rank $r\geq 1$, with compact singular value decomposition
\begin{equation*}
Z
=
U_Z\Sigma_ZV_Z^\top.
\end{equation*}
The canonical partial-isometry polar map is defined by
\begin{equation*}
\mathcal{P}(Z)
=
U_ZV_Z^\top
=
U_Z I_r V_Z^\top.
\end{equation*}
Hence, all nonzero singular values of $\mathcal{P}(Z)$ are equal to one:
\begin{equation*}
\sigma_i\!\left(\mathcal{P}(Z)\right)
=
1,
\qquad
i=1,\ldots,r.
\end{equation*}

\paragraph{Modulation before the polar map.}
Let $D\in\mathbb{R}_{>0}^{m\times n}$ be a positive element-wise gate, and define the modulated proxy
\begin{equation*}
Y
=
D\odot Z.
\end{equation*}
Let $q=\operatorname{rank}(Y)$ and write its compact SVD as
\begin{equation*}
Y
=
U_Y\Sigma_YV_Y^\top.
\end{equation*}
The polar map applied after modulation gives
\begin{equation*}
\mathcal{P}(D\odot Z)
=
\mathcal{P}(Y)
=
U_YV_Y^\top
=
U_Y I_q V_Y^\top.
\end{equation*}
Therefore,
\begin{equation*}
\sigma_i\!\left(\mathcal{P}(D\odot Z)\right)
=
1,
\qquad
i=1,\ldots,q.
\end{equation*}
Thus, although the element-wise gate may alter the rank and singular structure of the proxy, the subsequent polar map normalizes all nonzero singular values to one.
Applying a common update scale $s_{\mathrm{scale}}>0$ then gives
\begin{equation*}
\sigma_i\!\left(
s_{\mathrm{scale}}\mathcal{P}(D\odot Z)
\right)
=
s_{\mathrm{scale}},
\qquad
i=1,\ldots,q,
\end{equation*}
so all nonzero singular values remain equal.

\paragraph{Modulation after the polar map.}
The equal-nonzero-singular-value property is not generally preserved when a nonuniform element-wise gate is applied after the polar map.
Consider the square case $Z=I_n$ with $n\geq 2$, for which
\begin{equation*}
\mathcal{P}(I_n)
=
I_n.
\end{equation*}
Let $D\in\mathbb{R}_{>0}^{n\times n}$ have unequal diagonal entries $d_1,\ldots,d_n$.
Then
\begin{equation*}
D\odot\mathcal{P}(Z)
=
D\odot I_n
=
\operatorname{diag}(d_1,\ldots,d_n).
\end{equation*}

Its singular values are $d_1,\ldots,d_n$, up to ordering, and hence are not all equal.
Thus, nonuniform modulation after orthogonalization need not preserve equality among the nonzero singular values.

\paragraph{Finite-step Newton--Schulz approximation.}
In practice, Muon uses the $K$-step Newton--Schulz map $\operatorname{NS}_{K}$ rather than the exact polar map.
For the modulated proxy $Y=D\odot Z$ with rank $q$, define the approximation error
\begin{equation*}
\delta_K(Y)
=
\left\|
\operatorname{NS}_{K}(Y)
-
\mathcal{P}(Y)
\right\|_2.
\end{equation*}
The Weyl singular-value perturbation inequality provides a uniform bound for all $i=1,\ldots,q$:
\begin{equation*}
\left|
\sigma_i\!\left(\operatorname{NS}_{K}(Y)\right)
-
\sigma_i\!\left(\mathcal{P}(Y)\right)
\right|
\leq
\delta_K(Y).
\end{equation*}
Since every nonzero singular value of \(\mathcal{P}(Y)\) equals one, the preceding inequality directly yields
\begin{equation*}
\left|
\sigma_i\!\left(\operatorname{NS}_{K}(Y)\right)
-
1
\right|
\leq
\delta_K(Y),
\qquad
i=1,\ldots,q.
\end{equation*}
For each $i=1,\ldots,q$, scaling by $s_{\mathrm{scale}}>0$ gives
\begin{equation*}
\left|
\sigma_i\!\left(
s_{\mathrm{scale}}\operatorname{NS}_{K}(Y)
\right)
-
s_{\mathrm{scale}}
\right|
\leq
s_{\mathrm{scale}}\delta_K(Y).
\end{equation*}
Thus, operator-norm proximity to the exact polar factor bounds deviations of the nonzero singular values from the equal-singular-value profile.


Muon-NSR and Muon-VS modulate the update proxy with positive element-wise variance gates before Newton--Schulz orthogonalization.
Under the exact polar map, this ordering preserves the equal-nonzero-singular-value structure of the update; with finite-step Newton--Schulz, the perturbation bound above controls deviations from this idealized structure.
By contrast, Muon-NSR-Reshuffled applies a generally nonuniform element-wise gate after orthogonalization.
The preceding counterexample shows that even exact orthogonalization does not ensure preservation of the equal-nonzero-singular-value structure after such post-orthogonalization modulation.
This establishes a structural asymmetry between the two orderings and provides a geometric rationale for applying variance modulation before orthogonalization.

\begin{table*}[t]
\centering
\small
\setlength{\tabcolsep}{4.2pt}
\renewcommand{\arraystretch}{1.08}
\begin{tabular*}{\textwidth}{@{\extracolsep{\fill}}llcccc@{}}
\toprule
Model
& Optimizer
& Tokens/s (K) $\uparrow$
& ms/update $\downarrow$
& Peak GPU Mem. (GiB) $\downarrow$
& Opt. State Mem. (GiB) $\downarrow$ \\
\midrule
\multirow{3}{*}{Suite~A 130M}
& Muon      & 458.3 & 1144.00 & 15.41 & 0.37 \\
& Muon-NSR  & 447.6 & 1171.41 & 15.66 & 0.62 \\
& Muon-VS   & 450.4 & 1164.06 & 15.66 & 0.62 \\
\midrule
\multirow{3}{*}{Suite~A 300M}
& Muon      & 274.1 & 1912.60 & 17.55 & 0.75 \\
& Muon-NSR  & 270.8 & 1936.07 & 18.11 & 1.31 \\
& Muon-VS   & 271.6 & 1930.59 & 18.11 & 1.31 \\
\midrule
\multirow{3}{*}{Suite~A 1.2B}
& Muon      & 99.5 & 10535.07 & 19.28 & 2.62 \\
& Muon-NSR  & 99.6 & 10526.42 & 21.53 & 4.87 \\
& Muon-VS   & 99.7 & 10520.68 & 21.53 & 4.87 \\
\midrule
\multirow{3}{*}{Suite~B 210M}
& Muon      & 416.2 & 314.89  & 15.37 & 0.92 \\
& Muon-NSR  & 400.6 & 327.22  & 16.01 & 1.55 \\
& Muon-VS   & 408.2 & 321.08  & 16.03 & 1.55 \\
\midrule
\multirow{3}{*}{Suite~B 720M}
& Muon      & 207.3 & 4899.25 & 19.72 & 3.06 \\
& Muon-NSR  & 205.7 & 4939.20 & 22.02 & 5.36 \\
& Muon-VS   & 206.1 & 4929.15 & 22.01 & 5.36 \\
\midrule
\multirow{3}{*}{GPT-2 Small}
& Muon      & 935.0 & 525.70  & 20.51 & 0.69 \\
& Muon-NSR  & 927.0 & 530.24  & 20.55 & 0.73 \\
& Muon-VS   & 925.9 & 530.84  & 20.55 & 0.73 \\
\midrule
\multirow{3}{*}{GPT-2 Medium}
& Muon      & 305.1 & 1611.23 & 16.99 & 1.03 \\
& Muon-NSR  & 304.4 & 1614.65 & 17.13 & 1.17 \\
& Muon-VS   & 304.4 & 1614.46 & 17.13 & 1.17 \\
\bottomrule
\end{tabular*}
\caption{\textbf{Runtime and memory overhead on Suite~A, Suite~B, and GPT-2 models.}
We report padded-token throughput, average wall-clock time per optimizer update, peak GPU memory, and maximum per-rank optimizer-state memory.
Suite~A and GPT-2 rows are measured by microbenchmarks over 200 updates after 10 warmup updates; Suite~B rows use dedicated 200-update runs.
All measurements use the effective global batch size for each training configuration.
Tokens/s is in thousands, ms/update in milliseconds, and memory in GiB.}
\label{tab:optimizer_overhead_gpt_suitea_suiteb}
\end{table*}

\subsection{Muon-NSR-Reshuffled}
\label{sec:supp_algorithms}


For completeness, Algorithm~\ref{alg:muon_nsr_reshuffled} specifies Muon-NSR-Reshuffled, the post-orthogonalization variant used in the controlled ordering ablation.
Compared with Muon-NSR, this variant applies a corresponding coordinate-wise NSR attenuation at the post-orthogonalization stage, enabling a direct controlled assessment of the effect of modulation ordering while keeping all remaining optimization components fixed.

\paragraph{GPT-2 Small ordering ablation.}
To examine whether the benefit of pre-orthogonalization NSR modulation extends beyond Llama-style pretraining, we evaluate Muon-NSR-Reshuffled on GPT-2 Small (125M) under the \texttt{nanoGPT}-based setup described in Appendix~\ref{sec:supp_nanogpt_setup}.
As shown in Figure~\ref{fig:gpt2_125m_reshuffled}, the reshuffled variant achieves lower validation loss than AdamW and Muon, showing that post-orthogonalization NSR attenuation can still improve over these baselines.
However, it consistently underperforms Muon-NSR and remains above AdaMuon later in training.
Together with the Llama-130M ablation, this result provides empirical support for the structural argument in Appendix~\ref{app:formal_ordering} and favors applying NSR modulation before Newton--Schulz orthogonalization.

\section{Runtime and Memory Overhead}
\label{app:optimizer_overhead}

We quantify the runtime and memory overhead introduced by the additional variance buffer and element-wise modulation in Muon-NSR and Muon-VS on a single node with $8\times$ NVIDIA RTX~5090 GPUs.
We report padded-token throughput, average wall-clock time per optimizer update, peak GPU memory, and optimizer-state memory.
Measurements for Suite~A, Suite~B, and GPT-2 use their corresponding training configurations.
For Suite~A and GPT-2, each microbenchmark averages 200 measured optimizer updates following 10 warmup updates.
For Suite~B Llama-210M and Llama-720M, we report the average wall-clock time per optimizer update from dedicated 200-update measurement runs.
Token throughput is computed using padded tokens under fixed-length training.

Token throughput for optimizer $o$ is computed as
\begin{equation*}
\mathrm{Tokens/s}(o)
=
\frac{
B_{\mathrm{eff}} \cdot L
}{
\bar{t}_{\mathrm{update}}(o)
},
\end{equation*}
where $B_{\mathrm{eff}}$ is the effective global batch size in sequences per optimizer update, $L$ is the sequence length, and $\bar{t}_{\mathrm{update}}(o)$ is the average wall-clock time per optimizer update.
For Suite~A and GPT-2 microbenchmarks,
$B_{\mathrm{eff}}=B_{\mathrm{micro}}N_{\mathrm{gpu}}A_{\mathrm{grad}}$
under the corresponding training configuration; for Suite~B measurements, $B_{\mathrm{eff}}$ is the global batch size of the corresponding 200-update run.
Optimizer-state memory is reported as the maximum per-rank storage of optimizer-state tensors:
\begin{equation*}
M_{\mathrm{opt}}
=
\frac{1}{2^{30}}
\max_{r}
\sum_{s\in\mathcal{S}_{\mathrm{opt}}^{(r)}}
\operatorname{numel}(s)\cdot \operatorname{bytes}(s),
\end{equation*}
where $\mathcal{S}_{\mathrm{opt}}^{(r)}$ is the set of optimizer-state tensors stored on rank $r$, and $\operatorname{bytes}(s)$ denotes the number of bytes per element of tensor $s$.


As shown in Table~\ref{tab:optimizer_overhead_gpt_suitea_suiteb}, these single-node measurements characterize implementation overhead.
Across the Suite~A, Suite~B, and GPT-2 settings considered in this evaluation, Muon-NSR and Muon-VS exhibit only modest changes in per-update wall-clock time while consistently maintaining token throughput close to Muon.
This empirical behavior is consistent with the complexity analysis in Section~\ref{sec:method}, where the additional $O(mn)$ element-wise variance modulation is dominated by the $O(Kmn\min(m,n))$ Newton--Schulz computation for typical matrix-shaped Transformer parameter groups.

The additional variance buffer increases the optimizer-state memory and moderately increases peak GPU memory in these measurements.
The variance buffer is maintained only for matrix-shaped parameters in Transformer blocks using Muon updates, while embedding tables and vector-shaped parameters remain under AdamW and receive no additional variance state.
Thus, this additional state is confined exclusively to the Muon-optimized parameter groups.
The higher observed peak memory of GPT-2 Small relative to GPT-2 Medium stems from its larger per-GPU microbatch under a different gradient accumulation setting, which increases activation memory despite the smaller model.

\section{Small-Model Zero-Shot Downstream Evaluation}
\label{app:small_model_downstream}

To assess whether pretraining gains transfer to zero-shot downstream performance, we evaluate the Suite~A Llama-130M and Llama-300M checkpoints using EleutherAI's Language Model Evaluation Harness (\texttt{lm-eval-harness})~\citep{gao2021framework}.
We use \texttt{lm-eval-harness} v0.4.11 with Hugging Face model loading~\citep{wolf2020transformers}, \texttt{bfloat16} inference, automatic batch-size selection, and the bundled task configurations, including their default prompts, splits, post-processing rules, and metrics.
All evaluations are conducted in the zero-shot setting with \texttt{num\_fewshot=0}.

Given the relatively small model scales, we treat the downstream evaluation as a targeted sanity check rather than a comprehensive assessment of model capabilities.
Accordingly, we report results on a ten-task subset spanning commonsense reasoning, language modeling, reading comprehension, natural language inference, and truthfulness-oriented multiple-choice tasks.

For reproducibility, we report the task metadata versions used in our \texttt{lm-eval-harness} evaluation: v1.0 for \texttt{arc\_easy}, \texttt{hellaswag}, \texttt{lambada\_openai}, \texttt{piqa}, and \texttt{winogrande}; v2.0 for \texttt{race}, \texttt{record}, and \texttt{truthfulqa\_mc1}; v3.0 for \texttt{truthfulqa\_mc2}; and v0.0 for \texttt{sglue\_rte}.

\begin{table*}[t]
\centering
\small
\setlength{\tabcolsep}{4pt}
\renewcommand{\arraystretch}{1.08}
\begin{tabular*}{\textwidth}{@{\extracolsep{\fill}}lcccccc@{}}
\toprule
\multicolumn{7}{c}{\textbf{Llama-130M}} \\
\addlinespace[1pt]
Task & Muon & AdaMuon & NorMuon & Muon$^2$ & Muon-NSR & Muon-VS \\
\midrule
ARC-E~\citep{clark2018think} & 36.95 & 38.38 & 37.50 & 37.04 & 38.38 & 37.63 \\
HellaSwag~\citep{zellers2019hellaswag} & 28.08 & 28.39 & 28.49 & 28.08 & 28.93 & 28.60 \\
LAMBADA~\citep{paperno2016lambada} & 24.86 & 25.91 & 25.75 & 25.15 & 26.12 & 26.28 \\
PIQA~\citep{bisk2020piqa} & 59.85 & 60.39 & 59.85 & 60.50 & 59.63 & 59.19 \\
RACE~\citep{lai2017race} & 25.65 & 26.32 & 26.41 & 26.03 & 27.75 & 26.89 \\
ReCoRD~\citep{zhang2018record} & 52.90 & 52.72 & 53.90 & 53.46 & 54.57 & 53.89 \\
RTE~\citep{wang2019superglue} & 50.18 & 50.54 & 49.82 & 52.71 & 53.79 & 55.23 \\
TruthfulQA-MC1~\citep{lin2022truthfulqa} & 22.77 & 23.99 & 23.62 & 24.24 & 24.24 & 24.11 \\
TruthfulQA-MC2~\citep{lin2022truthfulqa} & 40.99 & 43.52 & 42.36 & 42.94 & 43.36 & 43.17 \\
WinoGrande~\citep{sakaguchi2020winogrande} & 51.30 & 51.07 & 51.78 & 50.91 & 51.54 & 51.46 \\
\multicolumn{1}{c}{Average}
& 39.35
& 40.12
& 39.95
& 40.11
& \textbf{40.83}
& \textbf{40.65} \\

\midrule
\multicolumn{7}{c}{\textbf{Llama-300M}} \\
\addlinespace[1pt]
Task & Muon & AdaMuon & NorMuon & Muon$^2$ & Muon-NSR & Muon-VS \\
\midrule
ARC-E~\citep{clark2018think} & 41.16 & 41.50 & 40.70 & 41.08 & 41.08 & 40.49 \\
HellaSwag~\citep{zellers2019hellaswag} & 31.11 & 31.40 & 30.99 & 31.45 & 31.61 & 31.25 \\
LAMBADA~\citep{paperno2016lambada} & 30.35 & 30.23 & 30.55 & 30.39 & 31.13 & 30.68 \\
PIQA~\citep{bisk2020piqa} & 61.86 & 62.40 & 61.64 & 62.30 & 61.70 & 62.46 \\
RACE~\citep{lai2017race} & 27.66 & 26.99 & 28.61 & 26.51 & 28.13 & 27.37 \\
ReCoRD~\citep{zhang2018record} & 58.57 & 59.11 & 59.23 & 58.25 & 59.53 & 59.81 \\
RTE~\citep{wang2019superglue} & 52.71 & 53.07 & 51.99 & 51.62 & 54.87 & 53.79 \\
TruthfulQA-MC1~\citep{lin2022truthfulqa} & 22.40 & 23.01 & 22.15 & 21.79 & 22.64 & 22.77 \\
TruthfulQA-MC2~\citep{lin2022truthfulqa} & 41.60 & 42.98 & 41.59 & 40.54 & 42.82 & 42.51 \\
WinoGrande~\citep{sakaguchi2020winogrande} & 51.93 & 51.38 & 50.83 & 49.88 & 50.12 & 51.62 \\
\multicolumn{1}{c}{Average}
& 41.94
& 42.21
& 41.83
& 41.38
& \textbf{42.36}
& \textbf{42.28} \\
\bottomrule
\end{tabular*}
\caption{\textbf{Small-model zero-shot transfer on Suite~A Llama checkpoints.}
Results are reported as percentages on the ten-task \texttt{lm-eval-harness} subset.
ARC-E, HellaSwag, and PIQA use \(\mathrm{acc}_{\mathrm{norm}}\), ReCoRD uses \(F_1\), and all other tasks use \(\mathrm{acc}\).
Average denotes the unweighted macro-average across the ten tasks.
AdaMuon, NorMuon, and Muon$^2$ serve as Muon-family baselines at both model scales.
Boldface in the Average row indicates the two highest macro-average scores at each model scale.
Higher values are better.}
\label{tab:small_model_zero_shot}
\end{table*}

Table~\ref{tab:small_model_zero_shot} summarizes the small-model zero-shot transfer results.
On Llama-130M, Muon-NSR and Muon-VS raise Muon's unweighted macro-average of \(39.35\) to \(40.83\) and \(40.65\), gains of \(+1.48\) and \(+1.30\) percentage points, respectively.
On Llama-300M, they raise it from \(41.94\) to \(42.36\) and \(42.28\), gains of \(+0.42\) and \(+0.34\) percentage points, respectively.
At both scales, the proposed variants outperform all evaluated Muon-family baselines in the unweighted macro-average across tasks.

Task-level gains are mixed, with several evaluated commonsense and physical-reasoning tasks remaining close to or slightly below the Muon baseline for one or both proposed variants.
Nevertheless, both Muon-NSR and Muon-VS improve the unweighted macro-average relative to all evaluated Muon-family baselines at both model scales.
Taken together with the pretraining results, these findings suggest that variance-adaptive Muon updates improve validation loss and convergence efficiency while preserving and, on average, improving zero-shot transfer in this small-model setting.

\end{document}